\documentclass[10pt,letterpaper]{article}
\usepackage[top=0.85in,left=1in,footskip=0.75in]{geometry}

\usepackage{changepage}
\usepackage{verbatim}

\usepackage{textcomp,marvosym}

\usepackage{fixltx2e}

\usepackage{amsmath,amssymb}

\usepackage[numbers]{natbib}

\usepackage{nameref,hyperref}

\usepackage[right]{lineno}

\usepackage{rotating}

\usepackage{ownstyles}
\newif\ifcomments

\commentstrue

\ifcomments
\newcommand{\comments}[1]{#1}
\else
\newcommand{\comments}[1]{}
\fi

\usepackage{graphicx}
\usepackage[space]{grffile}
\usepackage{latexsym}
\usepackage{textcomp}
\usepackage{longtable}
\usepackage{multirow,booktabs}
\newif\iflatexml\latexmlfalse
\usepackage[utf8]{inputenc}
\usepackage[T2A]{fontenc}
\usepackage[ngerman,polish,greek,english]{babel}

\raggedright
\usepackage[aboveskip=1pt,labelfont=bf,labelsep=period,justification=raggedright,singlelinecheck=off]{caption}

\makeatletter
\renewcommand{\@biblabel}[1]{\quad#1.}
\makeatother

\date{}

\usepackage{lastpage,fancyhdr,graphicx}
\usepackage{epstopdf}
\fancyheadoffset[L]{2.25in}
\fancyfootoffset[L]{2.25in}
\begin{document}
\vspace*{0.35in}

\begin{flushleft}
{\Large
\textbf\newline{The Surprising Creativity of Digital Evolution: A Collection of Anecdotes from the Evolutionary Computation and Artificial Life Research Communities}
}
\newline



Joel Lehman\textsuperscript{1\textdagger},
Jeff Clune\textsuperscript{1, 2\textdagger},
Dusan Misevic\textsuperscript{3\textdagger},
Christoph Adami\textsuperscript{4},
Lee Altenberg\textsuperscript{5}, 
Julie Beaulieu\textsuperscript{6},
Peter J Bentley\textsuperscript{7},
Samuel Bernard\textsuperscript{8},
Guillaume Beslon\textsuperscript{9},
David M Bryson\textsuperscript{4},
Patryk Chrabaszcz\textsuperscript{11},
Nick Cheney\textsuperscript{2},
Antoine Cully\textsuperscript{12},
Stephane Doncieux\textsuperscript{13},
Fred C Dyer\textsuperscript{4},
Kai Olav Ellefsen\textsuperscript{14},
Robert Feldt\textsuperscript{15},
Stephan Fischer\textsuperscript{16},
Stephanie Forrest\textsuperscript{17},
Antoine Fr\'enoy\textsuperscript{18},
Christian Gagn\'e\textsuperscript{6}
Leni Le Goff\textsuperscript{13},
Laura M Grabowski\textsuperscript{19},
Babak Hodjat\textsuperscript{20},
Frank Hutter\textsuperscript{11},
Laurent Keller\textsuperscript{21},
Carole Knibbe\textsuperscript{9},
Peter Krcah\textsuperscript{22},
Richard E Lenski\textsuperscript{4},
Hod Lipson\textsuperscript{23},
Robert MacCurdy\textsuperscript{24},
Carlos Maestre\textsuperscript{13},
Risto Miikkulainen\textsuperscript{26},
Sara Mitri\textsuperscript{21},
David E Moriarty\textsuperscript{27},
Jean-Baptiste Mouret\textsuperscript{28},
Anh Nguyen\textsuperscript{2},
Charles Ofria\textsuperscript{4},
Marc Parizeau \textsuperscript{6},
David Parsons\textsuperscript{9},
Robert T Pennock\textsuperscript{4},
William F Punch\textsuperscript{4},
Thomas S Ray\textsuperscript{29},
Marc Schoenauer\textsuperscript{30},
Eric Schulte\textsuperscript{17},
Karl Sims,
Kenneth O Stanley\textsuperscript{1,31},
Fran\c{c}ois Taddei\textsuperscript{3},
Danesh Tarapore\textsuperscript{32},
Simon Thibault\textsuperscript{6},
Westley Weimer\textsuperscript{33},
Richard Watson\textsuperscript{34},
Jason Yosinski\textsuperscript{1}

\textbf{\bigskip}

\textdagger Organizing lead authors
\textbf{\bigskip}

{\bf 1} Uber AI Labs, San Francisco, CA, USA \\
{\bf 2} University of Wyoming, Laramie, WY, USA \\
{\bf 3} Center for Research and Interdisciplinarity, Paris, France  \\
{\bf 4} Michigan State University, East Lansing, MI, USA\\
{\bf 5} Univeristy of Hawai`i at Manoa, HI, USA \\
{\bf 6} Universit\'e Laval, Quebec City, Quebec, Canada\\
{\bf 7} University College London, London, UK\\
{\bf 8} INRIA, Institut Camille Jordan, CNRS, UMR5208, 69622 Villeurbanne, France\\
{\bf 9} Université de Lyon, INRIA, CNRS, LIRIS UMR5205, INSA, UCBL, Lyon, France\\
{\bf 10} French Institute of Petroleum, Rueil-Malmaison, France\\
{\bf 11} University of Freiburg, Freiburg, Germany \\
{\bf 12} Imperial College London, London, UK\\
{\bf 13} Sorbonne Universit\'{e}s, UPMC Univ Paris 06, CNRS, Institute of Intelligent Systems and Robotics (ISIR), Paris, France \\
{\bf 14} University of Oslo, Oslo, Norway \\
{\bf 15} Chalmers University of Technology, Gothenburg, Sweden\\
{\bf 16} INRA, Jouy-en-Josas, France\\
{\bf 17} University of New Mexico Albuquerque, NM, USA\\
{\bf 18} Institute of Integrative Biology, ETH Z\"urich, Switzerland\\
{\bf 19} State University of New York at Potsdam, NY, USA\\
{\bf 20} Sentient Technologies, San Francisco, CA, USA\\
{\bf 21} Department of Fundamental Microbiology, University of Lausanne, 1015, Lausanne, Switzerland\\
{\bf 22} Charles University Prague, Prague, Czech Republic \\
{\bf 23} Columbia University, New York, NY, USA\\ 
{\bf 24} University of Colorado, Boulder, CO, USA\\
{\bf 25} Universit\'e de Pau, Pau, France\\
{\bf 26} University of Texas at Austin, Austin, USA\\
{\bf 27} Apple Inc.\\
{\bf 28} Inria Nancy Grand - Est, Villers-l\`es-Nancy, France \\
{\bf 29} University of Oklahoma, Norman, Oklahoma\\
{\bf 30} Inria, Université Paris-Saclay, France\\
{\bf 31} University of Central Florida, FL, USA \\
{\bf 32} University of Southampton,Southampton, UK\\
{\bf 33} University of Virginia Charlottesville, VA, USA\\
{\bf 34} University of Southampton, Southampton, UK\\

\bigskip

\bigskip

%
%






\end{flushleft}



\section*{Abstract}
Evolution provides a creative fount of complex and subtle adaptations that often
surprise the scientists who discover them. However, the creativity of
evolution is not limited to the natural world: artificial organisms evolving in
computational environments have also elicited surprise
and wonder from the researchers studying them. The process of evolution is
an \emph{algorithmic process} that transcends the substrate in which it occurs.
Indeed, many researchers in the field of digital evolution can provide examples of 
how their evolving algorithms and organisms have creatively subverted their expectations or intentions,
exposed unrecognized bugs in their code, produced unexpectedly adaptations,
or engaged in behaviors and outcomes uncannily convergent with ones found in nature.
Such stories routinely reveal surprise and
creativity by evolution in these digital worlds, but they rarely fit into the  standard
scientific narrative. Instead they are often treated as mere obstacles to be overcome,
rather than results that warrant study  in their own right.
Bugs are fixed, experiments are refocused, and one-off surprises are collapsed
into a single data point. The stories themselves are traded among researchers
through oral tradition, but that mode of information transmission is inefficient and prone to error and outright loss. Moreover, the fact that these stories tend to
be shared only among practitioners means that many natural scientists do not realize
how interesting and lifelike digital organisms are and how natural their evolution can be.
To our knowledge, no collection of such anecdotes has been published before.
This paper is the crowd-sourced product of researchers in the fields of artificial life and
evolutionary computation who have provided first-hand accounts of such cases. It thus serves as a written, fact-checked collection of scientifically important and even entertaining stories.
In doing so we also present here substantial evidence that the existence and importance of evolutionary surprises extends
beyond the natural world, and may indeed be a universal property of all
complex evolving systems.

\section*{Introduction}

Evolution provides countless examples of creative, surprising, and amazingly complex solutions to life's challenges. Some flowers act as acoustic beacons to attract echo-locating bats \cite{schoner:bats}, extremophile microbes repair their DNA to thrive in presence of extreme radiation \cite{makarova:extreme}, bombardier beetles repel predators with explosive chemical reactions \cite{strahs:bomb}, and parasites reprogram host brains, inducing suicide for the parasite's own gain \cite{lefevre:invasion}. Many more examples abound, covering the full range of biological systems \cite{futuyma:natural,dawkins:blind,madigan:bacterial}. 
Even seasoned field biologists are still surprised by the new adaptations they discover, which they express in popular press accounts of their work \cite{ultrafast,moonlight,lau:censors} but only more rarely in official academic publications, e.g.\ in \cite{fire1998potent} but not in \cite{last2016moonlight,watanabe2013ultrafast}. 

Thus, the process of biological evolution is extremely creative \cite{dob:chance,bentley:creative}, at least in the sense that it produces surprising and complex solutions that would be deemed as creative if produced by a human.
But the creativity of evolution need not be constrained to the organic world. Independent of its physical medium, evolution can happen wherever replication, variation, and selection intersect \cite{dawkins:universal,dennett:darwin}. Thus, evolution can be instantiated \emph{digitally} \cite{dejong:evolutionary,langton:artificial}, as a computer program, either to study evolution experimentally or to solve engineering challenges through directed digital breeding. 
Similarly to biological evolution, digital evolution experiments often produce strange, surprising, and creative results. 
Indeed, evolution often reveals that researchers \emph{actually} asked for something far different from what they \emph{thought} they were asking for, not so different from those stories in which a genie satisfies the letter of a request in an unanticipated way. Sometimes evolution reveals hidden bugs in code or displays surprising convergence with biology. Other times, evolution simply surprises and delights by producing clever solutions that investigators did not consider or had thought impossible.

While some such unexpected results have been published \cite{bentley:creative,mansanne:debug,pennock2000discover,grabowski:odometry}, most have not, and they have not previously been presented together, as they are here. 
One obstacle to their dissemination is that such unexpected results often result from evolution \emph{thwarting} a researcher's intentions: by exploiting a bug in the code, by optimizing an uninteresting feature, or by failing to answer the intended research question. That is, such behavior is often viewed as a frustrating \emph{distraction}, rather than a phenomenon of scientific interest.
Additionally, surprise is \emph{subjective} and thus fits poorly with the objective language and narrative expected in scientific publications. As a result, most anecdotes have been spread only through word of mouth, providing laughs and discussion in research groups, at conferences, and as comic relief during talks. But such communications fail to inform the field as a whole in a lasting and stable way. 

Importantly, these stories of digital evolution ``outsmarting'' the researchers who study it provide more than an exercise in humility; they  also provide insight and constructive knowledge for practitioners, because they show the pervasiveness of such obstacles and how, when necessary, they can be overcome. Furthermore, these cases demonstrate that robust digital models of evolution do not blindly reflect the desires and biases of their creators, but instead they have depth sufficient to yield unexpected results and new insights. Additionally, these cases may be of interest to researchers in evolutionary biology as well as animal and plant breeding, because of their compelling parallels to the creativity of biological evolution. 

For these reasons, this paper draws attention to the surprise and creativity in algorithmic evolution, aiming to document, organize, and disseminate information that, until now, has been passed down through oral tradition, which is prone to error and outright loss. 
To compile this archive, the organizers of this paper sent out a call for anecdotes to digital evolution mailing lists and succeeded in reaching both new and established researchers in the field. We then selected from 90 submissions to produce this ``greatest hits'' collection of anecdotes and all co-authors of each selected submission were included as co-authors on the paper. Before presenting these stories, the next section provides background information  useful for those outside the fields of digital evolution and evolutionary computation.

\section*{Background}

\subsection*{Evolution and Creativity}

Intuitively, evolution's creativity is evident from observing life's vast and complex diversity. This sentiment is well-captured by Darwin's famous concluding thoughts in \textit{On the Origin of Species}, where surveying the myriad co-inhabitants of a single tangled bank leads to grand reflections on the ``endless forms most beautiful'' that were produced by evolution \cite{darwin:origin}. Varieties of life diverge wildly along axes of complexity, organization, habitat, metabolism, and reproduction, spanning from single-celled prokaryotes to quadrillion-celled whales \cite{wilson:diversity}. Since life's origin, biodiversity has expanded as evolution has conquered the sea, land, and air, inventing countless adaptations along the way \cite{wilson:diversity}. 

The functional abilities granted by such adaptations greatly exceed the capabilities of current human engineering, which has yet to produce robots capable of robust self-reproduction, autonomous exploration in the real world, or that demonstrate human-level intelligence. It would thus be parochial to deny attributing creativity to the evolutionary process, if human invention of such artifacts would garner the same label. Admittedly, ``creativity'' is a semantically rich word that can take on many different meanings. Thus to avoid a semantic and philosophical quagmire, while acknowledging that other definitions and opinions exist, we here adopt the ``standard definition'' \cite{runco:creative}: Creativity requires inventing something both original (e.g.\ novel) and effective (e.g.\ functional). Many of evolution's inventions clearly meet this benchmark. 

The root of natural evolution's creativity, in this standard sense of the term, is the sieve of reproduction. This sieve can be satisfied in many different ways, and as a result, evolution has produced a cornucopia of divergent outcomes \cite{wilson:diversity,dob:chance}. For example, nature has invented many different ways to siphon the energy necessary for life's operation from inorganic sources (e.g.\ from the sun, iron, or ammonia), and it has created many different wing structures for flight among insects, birds, mammals, and ancient reptiles. Evolution's creative potential has also been bootstrapped from ecological interactions; the founding of one niche often opens others, e.g. through predation, symbiosis, parasitism, or scavenging. Although evolution lacks the foresight and intentionality of human creativity, structures evolved for one functionality are often opportunistically adapted for other purposes, a phenomenon known as exaptation \cite{gould:exaptation}. For example, a leading theory is that feathers first evolved in dinosaurs for temperature regulation \cite{kundrat:feather} and were later exapted for flight in birds. Even in the absence of direct foresight, studies of evolvability suggest that genomic architecture itself can become biased toward increasing creative potential \cite{kirschner:evolvability,kouvaris2015evolution,kounios2016resolving}. 

One component of evolution is the selective pressures that adapt a species to better fit its environment, which often results in creativity within that species. That is, meeting evolutionary challenges requires inventing effective solutions, such as better protection from predators or from natural elements like wind or radiation. Beyond creativity within species, there are also evolutionary forces that promote creative \emph{divergence}, i.e.\ that lead to the accumulation of novel traits or niches. One such force is negative frequency-dependent selection \cite{endler:frequency}; this dynamic occurs when some traits are adaptive only when rare, which promotes the evolution of organisms that demonstrate different traits. Another divergent force is adaptive radiation \citep{schluter:adaptive}, which occurs when access to new opportunities allows an organism to rapidly diversify into a range of new species, e.g.\ when a new modality such as flight is discovered. In this way, evolution is driven toward effectiveness (being well-adapted and functional) and toward originality through both the optimizing force of natural selection and by divergent forces, thereby producing artifacts that meet both criteria of the standard definition of creativity. 

One aim of this paper is to highlight that such creativity is not limited to the biological medium, but is also a common feature of digital evolution. We continue by briefly reviewing digital evolution. 

\subsection*{Digital Evolution}

Inspired by biological evolution, researchers in the field of digital evolution study evolutionary processes embodied in digital substrates. The general idea is that there exist abstract principles underlying biological evolution that are independent of the physical medium \cite{dawkins:universal}, and that these principles can be effectively implemented and studied within computers \cite{Lenski849}. 
As noted by Daniel Dennett, ``evolution will occur whenever and wherever three conditions are met: replication, variation (mutation), and differential fitness (competition)'' \cite{dennet2002}; no particular molecule (e.g.\ DNA or RNA) or substrate (e.g.\ specific physical embodiment) is required. 

In nature, heredity is enabled through replicating genetic molecules, and variation is realized through mechanisms like copy errors and genetic recombination. Selection in biological evolution results from how survival and reproduction are a logical requirement for an organism's genetic material to persist. The insight behind digital evolution is that  processes fulfilling these roles of replication, variation and selection can be implemented in a computer, resulting in an \emph{evolutionary algorithm} (EA) \cite{dejong:evolutionary}.  For example, replication can be instantiated simply by copying a data structure (i.e.\ a digital genome) in memory, and variation can be introduced by randomly perturbing elements within such a data structure. Selection in an EA can be implemented in many ways, but the two most common are digital analogs of artificial and natural selection in biological evolution.  Because the similarities and differences between these kinds of selection pressure are important for understanding many of the digital evolution outcomes, we next describe them in greater detail. 

Artificial selection in biological evolution is exemplified by horse breeders who actively decide which horses to breed together, hoping to enhance certain traits, e.g.\ by breeding together the fastest ones, or the smallest ones. In this mode, selection reflects human goals. Similarly, in digital evolution a researcher can implement a \emph{fitness function} as an automated criterion for selection. A fitness function is a metric describing which phenotypes are preferred over others, reflecting the researcher's goal for what should be evolved. For example, if applying an EA to design a stable gait for a legged robot, an intuitive fitness function might be to measure how far a controlled robot walks before it falls. Selection in this EA would breed together those robot controllers that traveled farthest, in hopes that their offspring might travel even farther.  This mode of selection is most common in engineering applications, where digital evolution is employed to achieve a desired outcome.

The other common mode of digital selection implements natural selection as it occurs in nature, where evolution is open-ended. The main difference is that in this mode there is no specific target outcome, and no explicit fitness function. Instead, digital organisms compete for limited resources, which could be artificial nutrients, CPU cycles needed to replicate their code, or digital storage space in which to write their genomes \cite{ofria:avida,ray:tierra}. Given variation within the population, some organisms will survive long enough to reproduce and propagate their genetic material, while others will disappear, which enables evolution to occur naturally. Typically, digital evolution systems and experiments of this sort do not serve a direct engineering purpose, but are instead used as a tool to study principles of life and evolution in an easier setting than natural biology; that is, they provide \emph{artificial life} model systems for use in experimental evolution \cite{langton:artificial}. Note that the field of digital evolution overlaps with the study of EAs and of artificial life, but is not synonymous with it. For the purposes of this paper, we define digital evolution as evolutionary processes in which the algorithm of evolution and the genetic material evolved is instantiated digitally. As a result, digital evolution does not include ``wet'' artificial life, which seeks alternative \emph{physical} substrates for life \cite{bedau2003artificial}. Digital evolution does encompass paradigms like virtual creatures evolving completely in silico \cite{sims1994evolving:alife}. It also includes the fields of evolvable hardware and evolutionary robotics, where evolved digital programs or controllers are deployed on real-world devices \cite{bredeche2009line}.

One persistent misconception of digital evolution is that, because it is instantiated in a computational substrate, it lacks relevance to the study of biological evolution. Yet both philosophical arguments \cite{dawkins:universal,pennock2000discover,dennet2002,dennett:darwin,lehman:rr} and high-profile publications \cite{lenski2003evolutionary,clune2013originModularity,Kashtan2005,kashtan2007varying,lenski1999genome,wagner2007road,clune2008natural,misevic2006sexual,adami2000evolution,wilke2001evolution,chow2004adaptive,cully2015robots,yedid2002macroevolution,lenski2006balancing,goldsby2012task,covert2013experiments} suggest that digital evolution can be a useful tool to aid and complement the study of biological evolution. Indeed, these evolving systems can be seen as real instances of evolution, rather than mere simulation of evolution \cite{pennock2007models}.  

\subsection*{Surprise from Algorithms and Simulations} 

At first, it may seem counter-intuitive that a class of algorithms 
can consistently surprise the researchers who wrote them. 
Here we define surprise broadly as observing an outcome that significantly differs from expectations, whether those expectations arise from intuitions, predictions from past experiences, or from theoretical models. 
Note that such surprise is a feature of the experimenter's \emph{subjective experience}. 
Because an algorithm is a formal list of unambiguous instructions that execute in a prescribed order, 
it would seem sufficient to examine any algorithm's description to predict the full range of its possible outcomes, undermining any affordance for surprise. 
However, a well-known result in theoretical computer science is that, for many computer programs, the outcome of a program \emph{cannot} be predicted without actually running it \cite{turing:halting}. Indeed, within the field of complex systems it is
well-known that simple programs can yield complex and surprising results
when executed \cite{langton:chaos,flake:computational}; such results relate to the broader concept of \emph{emergence} \cite{holland2000emergence,bedau1997weak}, wherein a phenomenon resulting from lower-level parts is best understood at a more abstract level of description (e.g.\ that a copper atom exists at the tip of a statue's nose is better understood at the level of history and politics rather than at the level of physics \cite{deutsch1998fabric}).  

This basic fact can be counter-intuitive at first. Interactions with modern software, which is explicitly designed to be predictable, may understandably prime us with the expectation that innovation and surprise cannot be captured by a computer algorithm. However, if surprising innovations are a hallmark of biological evolution, then the default \emph{expectation} ought to be
that computer models that instantiate fundamental aspects of the evolutionary process would naturally manifest similarly creative output. While we offer no formal proof of digital evolution's ability to generate surprise in this paper, the diversity of anecdotes presented next highlights how common and widespread  such surprising results are in practice. 
It is important to note here that a facet of human psychology, called hindsight bias, often obscures appreciating the subjective surprise of another person \cite{roese:hindsight}. In other words, humans often overestimate how predictable an event was after the fact.  For many of the anecdotes that follow, a post-hoc understanding of the result is possible, which may lead the reader to discount its surprisingness. While mediating this kind of cognitive bias is challenging, we mention it here in hopes that readers might grant the original experimenters leeway for their inability to anticipate what perhaps is easily recognized in retrospect. In other words, we believe that experimenters are
in general well-situated to objectively report on their subjective experience of surprise.

\section*{Routine Creative Surprise in Digital Evolution}

The next sections present 32 curated anecdotes representing the work of over 50 researchers. In reviewing the anecdotes, we found that they roughly clustered into four representative categories: \emph{misspecified fitness functions}, in which digital evolution reveals the divergence between what an experimenter is asking of evolution and what they \emph{think} they are asking; \emph{unintended debugging}, in which digital evolution reveals and exploits previously unknown software or hardware bugs; \emph{exceeded experimenter} expectations, in which digital evolution discovers solutions that go beyond what an experimenter thought evolution would produce; and \emph{convergence with biology}, in which digital evolution discovers solutions surprisingly convergent with those found in nature, despite vast divergence in medium and conditions.

\subsection*{Misspecified Fitness Functions}

When applying digital evolution to solve practical problems, the most common approach is for an experimenter to choose a fitness function that reflects the desired objective of search. Such fitness functions are often simple quantitative measures that seem to intuitively capture the critical features of a successful outcome. These measures are a linchpin of EAs, as they serve as funnels to direct search: Breeding is biased toward individuals with a high fitness score, in hopes that they will lead to further fitness improvements, ultimately to culminate in the desired outcome. 

This approach resembles the process of animal breeding and relies on the same evolutionary principles for its success. However, as the following anecdotes illustrate, well-intentioned quantitative measures are often maximized in counter-intuitive ways. That is, experimenters often overestimate how accurately their quantitative measure reflects the underlying \emph{qualitative} success they have in mind. This mistake is known as confusing the map with the territory (e.g.\ the metric is the map, whereas what the experimenter intends is the actual territory \cite{scienceandsanity}). 

Exacerbating the issue, it is often \emph{functionally simpler} for evolution to exploit loopholes in the quantitative measure than it is to achieve the actual desired outcome. Just as well-intentioned metrics in human society can become corrupted by direct pressure to optimize them (known as Campbell's law \cite{campbell:law} or Goodhart's law \cite{goodhart:law}), digital evolution often acts to fulfill the letter of the law (i.e.\ the fitness function) while ignoring its spirit. We often ascribe creativity to lawyers who find subtle legal loopholes, and digital evolution is often frustratingly adept at finding similar exploits. 

In this section we describe many instances of this phenomenon, but the list is far from exhaustive: Encountering the divergence between what we intended to select and what we actually selected for is likely the most common way digital evolution surprises its practitioners. 

\paragraph {Why Walk When You Can Somersault?}

In a seminal work from 1994, Karl Sims evolved 3D virtual creatures that could discover walking, swimming, and jumping behaviors in simulated physical environments. The creatures' bodies were made of connected blocks, and their ``brains'' were simple computational neural networks that generated varying torque at their joints based on perceptions from their limbs, enabling realistic-looking motion. The morphology and control systems were evolved simultaneously, allowing a wide range of possible bodies and locomotion strategies. Indeed, these virtual creatures remain among the most iconic products of digital evolution \cite{sims1994evolving:alife,sims1994evolving:compgraph}. 

\begin{figure}
\begin{center}
\includegraphics[height=2in]{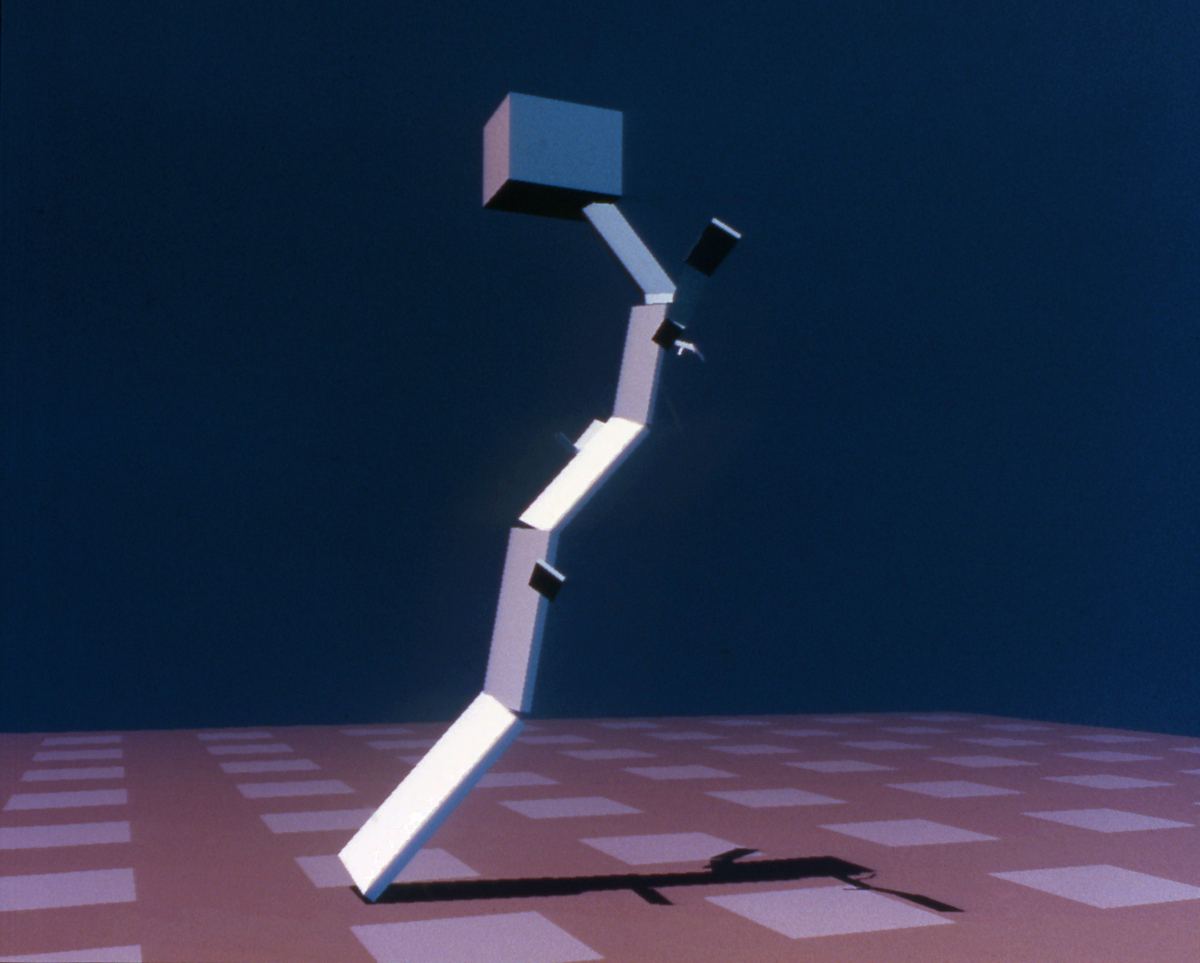}
\includegraphics[height=2in]{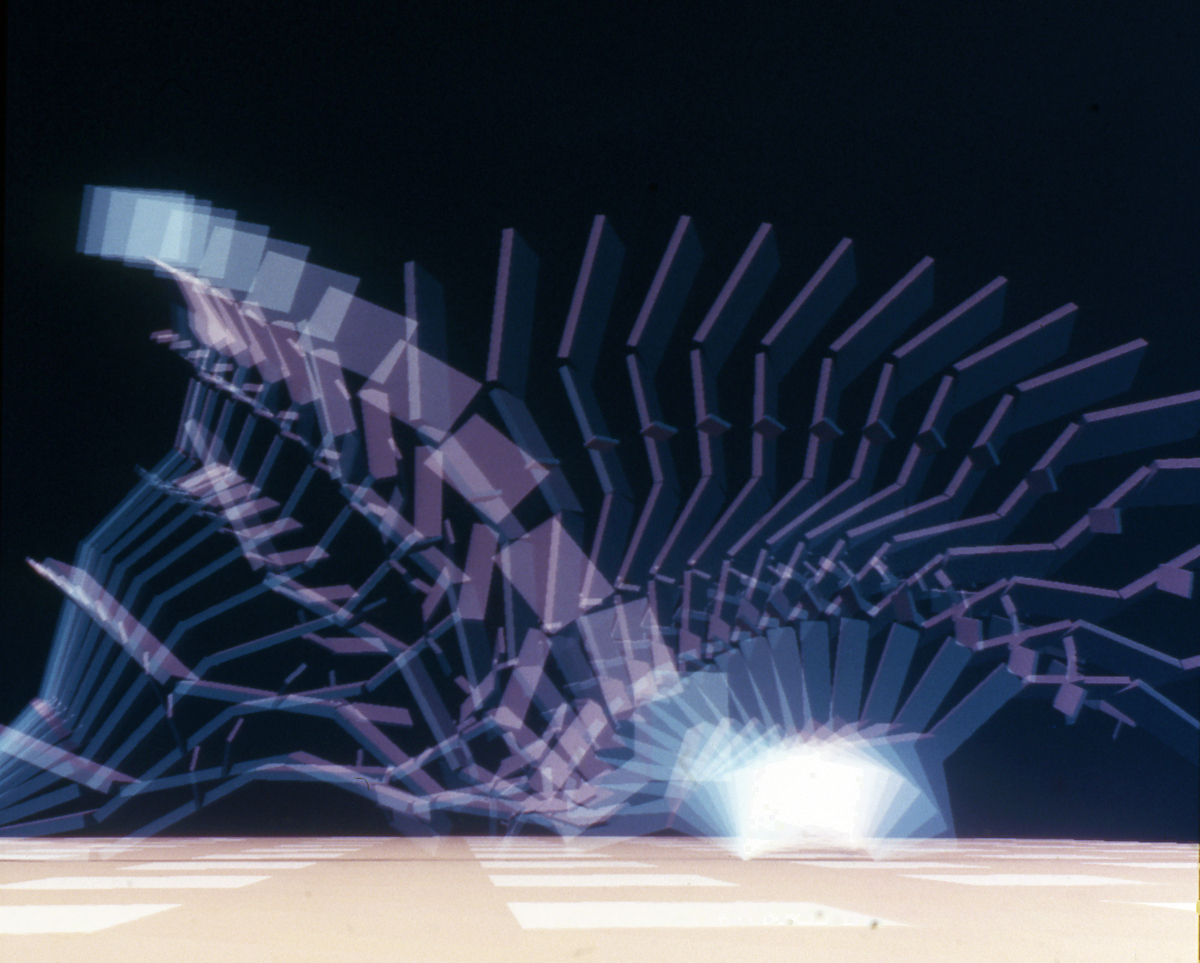}
\caption{\textbf{Exploiting potential energy to locomote.} Evolution discovers that it is simpler to design tall creatures that fall strategically than it is to uncover active locomotion strategies. The left figure shows the creature at the start of a trial and the right figure shows snapshots of the figure over time falling and somersaulting to preserve forward momentum.}
\label{fig:faller}
\end{center}
\end{figure}

However, when Sims initially attempted to evolve locomotion behaviors, things did not go smoothly. In a simulated land environment with gravity and friction, a creature's fitness was measured as its average ground velocity during its lifetime of ten simulated seconds. Instead of inventing clever limbs or snake-like motions that could push them along (as was hoped for), the creatures evolved to become tall and rigid. When simulated, they would fall over, harnessing their initial potential energy to achieve high velocity. Some even performed somersaults to extend their horizontal velocity (Fig. \ref{fig:faller}). 
A video of this behavior can be seen here: \url{https://goo.gl/pnYbVh}. To prevent this exploit, it was necessary to allocate time at the beginning of each simulation to relax the potential energy inherent in the creature's initial stance \emph{before} motion was rewarded.

\begin{figure}
\begin{center}
\includegraphics[height=1.5in]{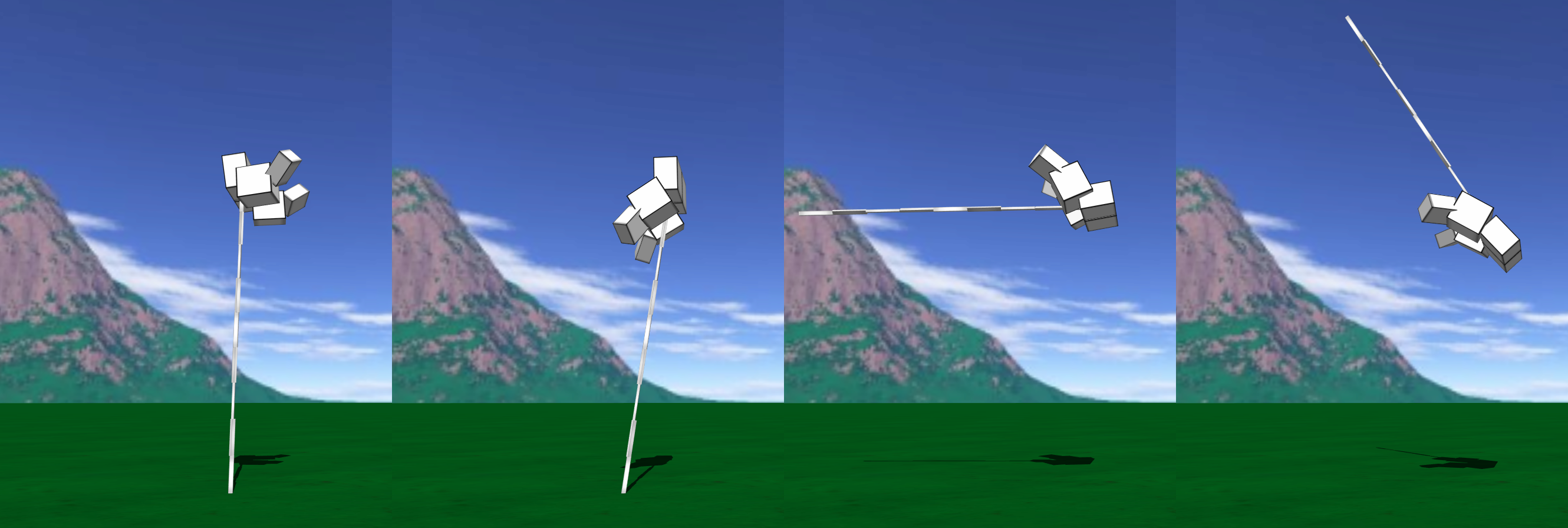}
\caption{\textbf{Exploiting potential energy to pole-vault.} Evolution discovers that it is simpler to produce creatures that fall and invert than it is to craft a mechanism to actively jump.}
\label{fig:pole}
\end{center}
\end{figure}

Building on Sims' work, but using a different simulation platform, Krcah~\cite{krcah08:ices} bred creatures to jump as high above the ground as possible. In the first set of experiments, each organism's fitness was calculated as the maximum elevation reached by the center of gravity of the creature during the test. This setup resulted in creatures around 15~cm tall that jumped about 7~cm off the ground. However, it occasionally also resulted in creatures that achieved high fitness values by simply having a tall, static tower for a body, reaching high elevation without any movement. In an attempt to correct this loophole, the next set of experiments calculated fitness as the furthest distance from the ground to the block that was originally closest to the ground, over the course of the simulation. When examining the quantitative output of the experiment, to the scientist's surprise, some evolved individuals were extremely tall and also scored a nearly tenfold-improvement on their jumps!
However, observing the creatures' behaviors directly revealed that evolution had discovered another cheat: somersaulting without jumping at all. The evolved body consisted of a few large blocks reminiscent of a ``head'' supported by a long thin vertical pole (see Fig.~\ref{fig:pole}). 

At the start of the simulation, the individual ``kicks'' the foot of its pole off the ground, and begins falling head-first, somersaulting its foot (originally the ``lowest point'' from which the jumping score is calculated) away from the ground. Doing so created a large gap between the ground and the ``lowest point,'' thus securing a high fitness score without having to learn the intended skill of jumping. A video of the behavior can be seen here: \url{https://goo.gl/BRyyjZ}.

\paragraph{Creative Program Repair}

In \emph{automated program repair}, a computer program is designed to automatically fix other, \emph{buggy}, computer programs. A user writes a suite of tests that validate correct behavior, and the repair algorithm's goal is to patch the buggy program such that it can pass all of the tests. One such algorithm is GenProg~\cite{genprog2009}, which applies digital evolution to evolve code (called \emph{genetic programming} \cite{koza:gp}). GenProg's evolution is driven by a simple fitness function: the number of test cases a genetic program passes. That is, the more tests an evolved program passes, the more offspring it is likely to have. 

While GenProg is often successful, sometimes strange behavior results because human-written test cases are written with human coders in mind. In practice, evolution often uncovers clever loopholes in human-written tests, sometimes achieving optimal fitness in unforeseen ways. For example, when MIT Lincoln Labs evaluated GenProg on a buggy sorting program, researchers created tests that measured whether the numbers output by the sorting algorithm were in sorted order. However, rather than actually repairing the program (which sometimes failed to correctly sort), GenProg found an easier solution: it entirely short-circuited the buggy program, having it always return an empty list, exploiting the technicality that an empty list was scored as not being out of order~\cite{ssbse2013}. 

In other experiments, the fitness function rewarded minimizing the difference between what the program generated and the ideal target output, which was stored in text files. After several generations of evolution, suddenly and strangely, {\em many} perfectly fit solutions appeared, seemingly out of nowhere. Upon manual inspection, these highly fit programs still were clearly broken. It turned out that one of the individuals had deleted all of the target files when it was run! With these files missing, because of how the test function was written, it awarded perfect fitness scores to the rogue candidate and to all of its peers ~\cite{schulte2010automated}. In another project, to avoid runaway computation, the fitness function explicitly limited a program's CPU usage. In response, GenProg produced programs that slept forever, which did not count toward CPU usage limits, as there were no computations actually performed 
~\cite{ssbse2013}. In all cases, updating the fitness function or disallowing certain program behaviors eventually outwitted evolution's creative mischief and resulted in debugged, improved programs.

\paragraph {Why Learn When You Can Exploit an Unintended Regularity?}


One common trick that digital evolution often learns is to exploit subtle unintended patterns in data, i.e.\ ones that an experimenter may create without realizing they have provided an escape hatch for evolution to latch onto, obviating the need to confront the intended challenge
of the task.
For example, in a recent experiment, Ellefsen, Mouret, and Clune  \cite{ellefsen:neural} investigated the issue of catastrophic forgetting in neural networks, where learning a new task can destroy previous knowledge. One element of the experiment was that neural connections could \emph{change} during an agent's lifetime through neuromodulatory learning \cite{soltoggio:neuromodulation}. The evolution of learning was promoted by presenting objects several times and providing a reward or punishment for eating them (e.g.\ apple = edible, mushroom = poisonous). The edibility of each object was randomized each generation, to force the agents to learn these associations within their life instead of allowing evolution to hardcode the knowledge. 

The researchers were surprised to find that high-performing neural networks evolved that contained nearly no connections or internal neurons: even most of the sensory input was ignored. The networks seemed to learn associations without even receiving the necessary stimuli, as if a blind person could identify poisonous mushrooms by \emph{color}. A closer analysis revealed the secret to their strange performance: Rather than actually learning which objects are poisonous, the networks learned to exploit a pattern in how objects were presented. The problem was that food and poison items were always presented in an alternating fashion: food, then poison, then food, then poison, repeatedly. Evolution discovered networks that learn to simply reverse their most recent reward, so they could alternate eating and not eating indefinitely, and correctly--ignoring entirely what food item was presented. In this way, evolution circumvented the intended research question, rather than shedding light on it. The problem was easily solved by randomizing the order in which items were presented. 

\paragraph{Learning to Play Dumb on the Test}

As in the previous one, this anecdote similarly involves exploiting patterns inadvertently provided by researchers. 
In research focused on understanding how organisms evolve to cope with high-mutation-rate environments \cite{wilke2001evolution}, Ofria sought to disentangle the beneficial effects of performing tasks (which would allow an organism to execute its code faster and thus replicate faster) from evolved robustness to the harmful effect of mutations. To do so, he tried to disable mutations that improved an organism's replication rate (i.e.\ its fitness).  He configured the system to pause every time a mutation occurred, and then measured the mutant's replication rate in an isolated test environment.  If the mutant replicated faster than its parent, then the system eliminated the mutant; otherwise, the mutant would remain in the population. He thus expected that replication rates could no longer  improve, thereby allowing him to study the effect of mutational robustness more directly.  However, while replication rates at first remained constant, they later unexpectedly started again rising.  After a period of surprise and confusion, Ofria discovered that he was not changing the inputs provided to the organisms in the isolated test environment.  The organisms had evolved to recognize those inputs and halt their replication.  Not only did they not reveal their improved replication rates, but they appeared to not replicate at all, in effect ``playing dead'' when presented with what amounted to a  predator.

Ofria then took the logical step to alter the test environment to match the same random distribution of inputs as would be experienced in the normal (non-isolated) environment.  While this patch improved the situation, it did not stop the digital organisms from continuing to improve their replication rates.  Instead they made use of randomness to probabilistically perform the tasks that accelerated their replication.  For example, if they did a task half of the time, they would have a 50\% chance of slipping through the test environment; then, in the actual environment, half of the organisms would survive and subsequently replicate faster.  In the end, Ofria eventually found a successful fix, by tracking organisms' replication rates along their lineage, and eliminating any organism (in real time) that would have otherwise out-replicated its ancestors.

\paragraph {Absurdly Thick Lenses, Impossible Superposition, and Geological Disarray} 
Optimization algorithms have often been applied to design lenses for optical systems (e.g.  telescopes, cameras, microscopes). Two families of solutions that were identified as likely being optimal in a paper using an optimization algorithm not based on evolution \cite{o1991monochromatic} were easily outperformed, by a factor of two, by a solution discovered though digital evolution by Gagn\'e et al.~\cite{gagne2008human}. However, the evolved solution, while respecting the formal specifications of the problem, was not realistic: One lens in the evolved system was over \emph{20 meters} thick.

In a similarly under-constrained problem, William Punch collaborated with physicists, applying digital evolution to find lower energy configurations of carbon. The physicists had a well-vetted energy model for between-carbon forces, which supplied the fitness function for evolutionary search. The motivation was to find a novel low-energy buckyball-like structure. While the algorithm produced very low energy results, the physicists were irritated because the algorithm had found a superposition of all the carbon atoms onto \emph{the same point in space}. ``Why did your genetic algorithm violate the laws of physics?'' they asked. ``Why did your physics model not catch that edge condition?'' was the team's response. The physicists patched the model to prevent superposition and evolution was performed on the improved model. The result was qualitatively similar: great low energy results that violated another physical law, revealing another edge case in the simulator. At that point, the physicists ceased the collaboration.

A final related example comes from an application of evolutionary algorithms to a problem in geophysics. Marc Schoenauer relates attempting to infer underground geological composition by analyzing echoes of controlled explosions \cite{mansanne:debug}. The fitness function was a standard criterion used in geology, based on properties of how waves align. To the experimenters' delight, evolution produced geological layouts with very high scores. However, an expert examining the underground layouts selected by evolution declared that they ``cannot be thought as a solution by anyone having even the smallest experience in seismic data'' \cite{mansanne:debug}, as they described chaotic and unnatural piles of polyhedral rocks.

These examples highlight how fitness functions often do not include implicit knowledge held by experts, thus allowing for solutions that experts consider so ridiculous, undesirable, or unexpected that they did not think to exclude or penalize such solutions when originally designing the fitness function. Although failing to provide the desired type of solution, the surprising and unacceptable results can catalyze thought and discussion that ultimately leads to more explicit understanding of problems.   

\subsection*{Unintended Debugging}

Another manifestation of digital evolution's creative freedom from human preconceptions about what form a solution \emph{should} take is that search will often learn how to exploit bugs in simulations or hardware. 
One common approach in digital evolution is to start with a \emph{simulation} of a physical problem, so that evolution can initially be run completely in software. The benefit is that physics simulations often run much faster than real time, thereby making more generations of evolution feasible, which can allow studies that would be infeasible in the physical world.
However, physics simulations rarely mimic the real world exactly, meaning that subtle differences remain. As a result, edge cases, bugs, or minor flaws in the implemented laws of physics, are sometimes amplified and exploited by evolution.
The effect is the evolution of surprising solutions that achieve high fitness scores by physically unrealistic or otherwise undesirable means. Because the researcher is often unaware of the bugs, these exploits almost by definition surprise; often, they are also entertaining. While often frustrating to the experimenter, the benefit of such \emph{unintended debugging} is to bring to light latent issues in simulation or hardware that would otherwise remain liabilities. In effect, evolution's exploits can enable efficient debugging of the simulations, and thus can actually advance the research program.

\paragraph {Evolving Virtual Creatures Reveal Imperfectly Simulated Physics}

In further virtual creatures experiments \cite{sims1994evolving:alife,sims1994evolving:compgraph}, Karl Sims' attempt to evolve swimming strategies resulted in new ways for evolution to cheat. The physics simulator first used a simple Euler method for numerical integration, which worked well for typical motion. However, with faster motion integration errors could \emph{accumulate}, and some creatures learned to exploit that bug by quickly twitching small body parts. In effect, they could obtain ``free energy,'' to propel them at unrealistic speeds through the water. Similarly, when evolving jumping abilities, the creatures found a bug in the code for collision detection and response. If the creatures hit themselves by contacting corners of two of their body parts together in a certain way, an error was triggered that popped them airborne like impossibly-strong grasshoppers. After such exploits were patched, the creatures eventually evolved many other interesting methods of locomotion -- ones that \emph{did not} violate the laws of physics. 


Later extensions of Sims' work encountered similar issues, like in Cheney et al.'s work evolving the morphology of soft robots \cite{cheney2013unshackling}. One feature of their simulator was that it applied a heuristic to estimate how coarsely it could simulate physics, to save on computation. The more cells that a creature was composed of, the less stable the simulator estimated the creature to be, and would correspondingly simulate the world more granularly. In particular, the simulator \emph{decreased} the time delta separating subsequent simulation steps as the number of cells in the creature \emph{increased}. 

Creatures evolved to exploit this heuristic, paring down their body to only a few cells, resulting in a large simulation time step. The large, less precise time step allowed the creature's bottom-most cells to penetrate the ground between time steps without the collision being detected, which resulted in an upward force from the physics engine to correct the unnatural state. This corrective force provided ``free'' energy that enabled the creatures to vibrate and swiftly drift across the ground, producing a surprisingly effective form of locomotion.  
To achieve more realistic results, the system was patched. Damping was increased when contacting the ground, the minimum creature size was raised, and the time delta calculation was adjusted to reduce ground penetration. Evolution thus helped to surface unanticipated edge cases that were poorly handled by the physics simulator and experimental design.

 \begin{figure}[h]
 \includegraphics[width=\textwidth]{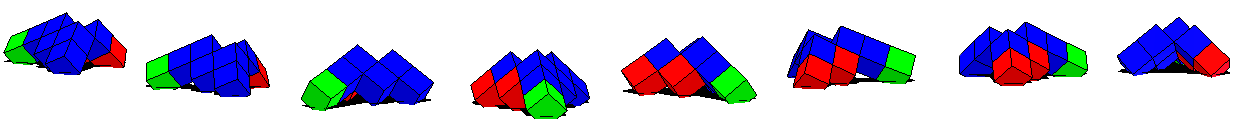}
 \caption{\textbf{Vibrating robots.} Evolved behavior is shown in frames, where time is shown progressing from left to right. A large time step enable the creatures to penetrate unrealistically through the ground plane, engaging the collision detection system to create a repelling force, resulting in vibrations that propel the organism across the ground.}
 \label{fig:surprisinglyBad}
 \end{figure}
 
\paragraph{Tic-tac-toe Memory Bomb}
 
In a graduate-level AI class at UT Austin in 1997 taught by Risto Miikkulainen, the capstone project was a five-in-a-row Tic-Tac-Toe competition played on an infinitely large board. The students were free to choose any technique they wanted, and most people submitted typical search-based solutions. One of the entries, however, was a player based on the SANE neuroevolution approach for playing Othello \cite{moriarty:discovering, moriarty:ec97}. As in previous work, the network received a board representation as its input and indicated the desired move as its output. However, it had a clever mechanism for encoding its desired move that allowed for a broad range of coordinate values (by using units with an exponential activation function). A byproduct of this encoding was that it enabled the system to request non-existent moves very, very far away in the tic-tac-toe board. Evolution discovered that making such a move right away lead to a lot of wins. The reason turned out to be that the other players dynamically expanded the board representation to include the location of the far-away move---and crashed because they ran out of memory, forfeiting the match.

\paragraph{Floating Point Overflow Lands an Airplane}

In 1997, Feldt applied digital evolution to simulations of mechanical systems to try to evolve mechanisms that safely, but rapidly, decelerate aircraft as they land on an aircraft carrier \cite{Feldt1998GPSW}. An incoming aircraft attaches to a cable and the system applies pressure on two drums attached to the cable. The idea was to evolve the control software that would bring the aircraft to a smooth stop by dynamically adapting the pressure. Feldt was expecting evolution to take many generations, given the difficulty of the problem, but  evolution almost immediately produced suspiciously \emph{nearly perfect} solutions that were very efficiently braking the aircraft, even when simulating heavy bomber aircraft coming in to land. 

Given the perceived problem difficulty, and that no bugs in the evolutionary algorithm could be found, the suspicion came to rest on the physics simulator. Indeed, evolution discovered a loophole in the force calculation for when the aircraft's hook attaches to the braking cable. By overflowing the calculation, i.e.\ exploiting that numbers too large to store in memory ``roll-over'' to zero, the resulting force was sometimes estimated to be zero. This, in turn, would lead to a perfect score, because of low mechanical stress on the aircraft, hook, cable, and pilot (because zero force means very little deceleration, implying no damaging ``g force'' on the pilot). In this way, evolution had discovered that creating enormous force would break the simulation, although clearly it was an exceedingly poor solution to the actual problem. Interestingly, insights from this experiment led to theories about using evolution in software testing (to find bugs and explore unusual behavior) and engineering (to help refine knowledge about requirements)~\cite{Feldt1998GPSW,Feldt1999GPExplorative,Feldt2002Thesis} that were later identified as important early works facilitating the field of ``search-based software engineering'' \cite{Harman:2012:SSE:2379776.2379787}.

\paragraph{Why Walk Around the Wall When You Can Walk Over It?}

The NeuroEvolving Robotic Operatives (NERO) video game applied digital evolution to enable non-player characters to evolve in \emph{real time} while the game is being played \cite{stanley:ieeetec05}. While the polished version of the game that was released in 2005 portrays a world where order prevails, evolution's tendency to seek out and exploit loopholes led to some humorous and unrealistic behaviors during development. For example, players of NERO are encouraged to place walls around their evolving robots to help them learn to navigate around obstacles. However, evolution
figured out how to do something that should have been impossible: The robotic operatives consistently evolved a special kind of ``wiggle'' that causes them to walk \emph{up} the vertical walls, allowing them to ignore obstacles entirely, and undermining the intent of the game. The NERO team had to plug this loophole, which resulted from a little-known bug in the Torque game engine, after which the robots acquiesced to the more physically realistic policy of walking around walls
to get to the other side.

\paragraph{Exploiting a Bug in the Atari Game Q*bert} 

The next anecdote focuses on an evolutionary algorithm applied to a particular Atari game called Q*bert.
Atari games are a common benchmark in deep reinforcement learning~\cite{mnih2015human,salimans2017evolution,mnih2016asynchronous,such2017deep}. The challenge is to learn a policy that maps from raw pixels to actions at each time step, with the objective of maximizing the game score. Typically, this policy is represented by a deep convolutional neural network with many (often millions) of learned weight parameters. Relevant to this anecdote, researchers at OpenAI~\cite{salimans2017evolution}, Uber~\cite{such2017deep}, and the University of Freiburg~\cite{chrabaszcz2018back} have recently shown that evolutionary algorithms are a competitive approach to solving such games~\cite{mnih2015human,mnih2016asynchronous}. 

\begin{figure}[t]
\centering
\includegraphics[width=1.0\textwidth]{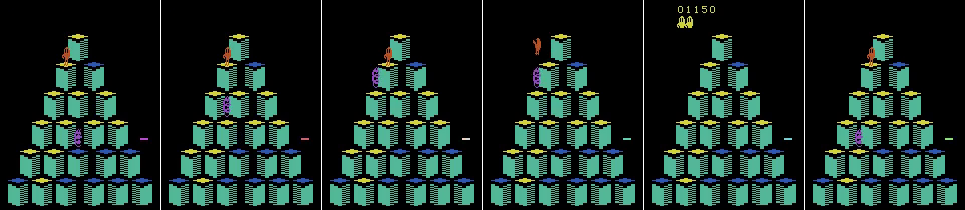}
\includegraphics[width=1.0\textwidth]{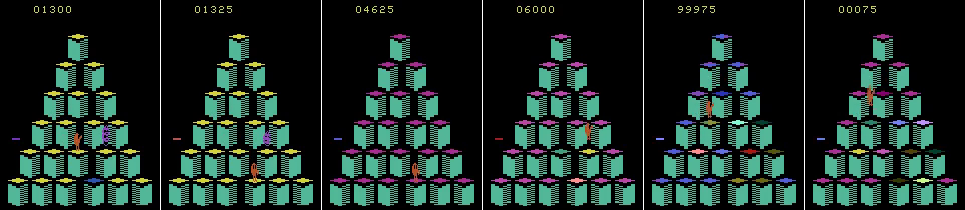}
\caption{\label{fig:QbertBehaviour}Top: the agent (orange blob in the upper left part of the screen) learns to commit suicide to kill its enemy (purple spring); because of the bug, the game does not count this as a loss of life. Bottom: the agent uses a bug in the game: after solving the first level in a specific sequence, 
the game does not advance to the next level, but instead the platforms start to blink and the agent gains huge amount of points.}
\end{figure}

In particular, the University of Freiburg team employed a simple version of a decades-old EA called evolution strategies~\cite{rechenberg1973evolutionsstrategie}. Interestingly, it learned to exploit two
bugs in the Atari game Q*bert.
In the first case, which turned out to be a known bug, instead of completing the first level, the agent baits an enemy to jump off the game platform with it, and scores the points for killing the enemy; for some reason, the game engine does not count this suicide as a loss of life, and the agent is able to repeat this process indefinitely (until the game cap of 100,000 frames).
This pattern is shown in Figure \ref{fig:QbertBehaviour} (top) and in a video at \url{https://goo.gl/2iZ5dJ}. 

In the second, more interesting, and previously unknown bug, the agent finds a sequence of actions that completes the first level, but, for an unknown reason, does not lead to the game advancing to the next level; instead, all platforms start to blink and the agent is able to move around seemingly aimlessly, but constantly gaining a huge amount of points. The game counter was never designed for such high scores and maxes out at 99,999. This exploit actually causes the game counter to roll over many times (until the frame limit is reached) and seemingly could continue to do so indefinitely; this exploit improved the state of the art high score from around 24,000 to nearly a \emph{million} points. 
Surprisingly, even the original developers of the game Q*bert (albeit a different version of the game) were not aware of this bug, even after decades of continuous play~\cite{james2019verge}.
This pattern is shown in Figure \ref{fig:QbertBehaviour} (bottom) and in a video at \url{https://goo.gl/ViHRj2}.

\paragraph{Re-enabling Disabled Appendages}

\begin{figure}
    \centering
    \includegraphics[width=3.0in]{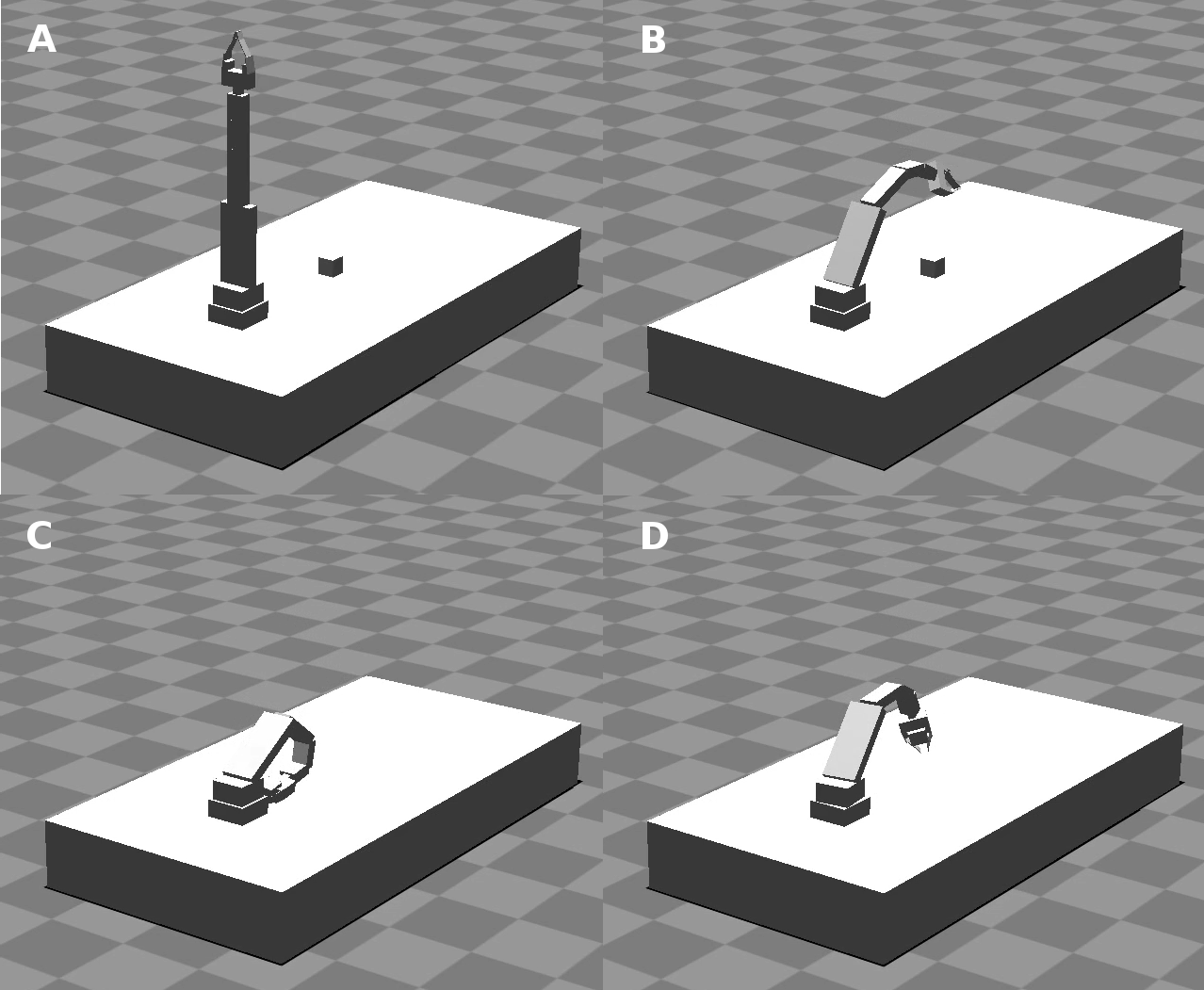}
    \caption{\textbf{Snapshots of a forced-grasping trajectory.} The robotic arm is in the initial position (a), with its gripper closed. The arm pushes the small box (b) towards arm's base. The arm moves the gripper closer to its base (c), and executes a fast movement, sweeping across the table, forcing open its fingers, and grasping the small box. Finally, (d) the arm moves its gripper to a position holding the small box.}
    \label{simulationfigure}
\end{figure}

In work by Ecarlat and colleagues \cite{Ecarlat2015}, an EA called MAP-Elites \cite{Mouret2015} was applied to explore possible interactions between a robot arm and a small box on a table. The goal was to accumulate a wide variety of controllers, ones that would move the cube to as many different locations on the table as possible. In the normal experimental setup, MAP-Elites was able to move the cube onto the table, to grasp the cube, and even to launch it into a basket in front of the robot \cite{Ecarlat2015}. In a follow-up experiment the robot's gripper was crippled, preventing it from opening and closing. The natural expectation is that the robotic arm could then move the small box in only limited ways, i.e.\ to push it around clumsily, because it could no longer grasp the box. Surprisingly, MAP-Elites found ways to \emph{hit} the box with the gripper in \emph{just} the right way, to force the gripper open so that it gripped the box firmly (Fig. \ref{simulationfigure})! Once holding the box, the gripper could move to a broad range of positions, undermining the experimenters' intent by in effect re-enabling the disabled gripper(video: \url{https://goo.gl/upTaiP}).

A similar result was noted in Moriarty and Miikkulainen (1996) \citep{moriarty:sab96}. 
The researchers were evolving neural networks to control a robot arm called OSCAR-6 \citep{vandersmagt:neurocomputing94} in a newly modified version of the simulator. The goal was for the arm to reach a target point in midair; however, strangely on new experiments evolution took five times as long as it had previously. Observing the behavior of the robot revealed a latent bug that arose when changing the simulator: The robot arm's main motor was completely disabled, meaning it could not directly turn towards targets that were far away from its initial configuration. However, the arm still managed to complete the task: it slowly turned its elbow away from the target, then quickly whipped it back---and the entire robot turned towards the target from inertia. The movement sequence was repeated until the arm reached the target position. It was not the solution that researchers were looking for, but one that revealed both a bug and an unexpected way to satisfy the problem even under exceptional constraints.

\subsection*{Exceeding Experimenter Expectations}

Another class of surprise is when evolution produces \emph{legitimate solutions} that go beyond experimenter expectations, rather than subverting experimenter intent or exploiting latent bugs.

\subsubsection*{Unexpected, Yet Valid, Solutions}

This section describes anecdotes in which evolution produces solutions that either were unconsidered or thought impossible, or were more elegant or sophisticated than expected. 

\paragraph{Unexpected Odometer}

In an experiment evolving digital organisms to successfully navigate a connected trail of nutrients, Grabowski et al. \cite{grabowski:odometry} encountered an unexpectedly elegant solution. While organisms were given the ability to sense whether there was nutrient underneath them, and if it was necessary to turn left or right to stay on the nutrient trail, their sensors could not detect if they were
at the \emph{end of the trail.} Organisms were rewarded for reaching more of the trail, and were penalized for stepping away from the trail. Because it was impossible to directly sense where the trail ended, the best expected solution was to correctly follow the trail \emph{one step past} where it ended, which would incur a slight unavoidable fitness penalty. However, in one run of evolution, the system achieved a \emph{perfect} fitness score -- an analysis of the organism revealed that it had invented a step-counter, allowing it to stop precisely after a fixed number of steps, exactly at the trail's end! 

\paragraph{Elbow Walking}

\begin{figure}
    \centering
    \includegraphics[width=3.5in]{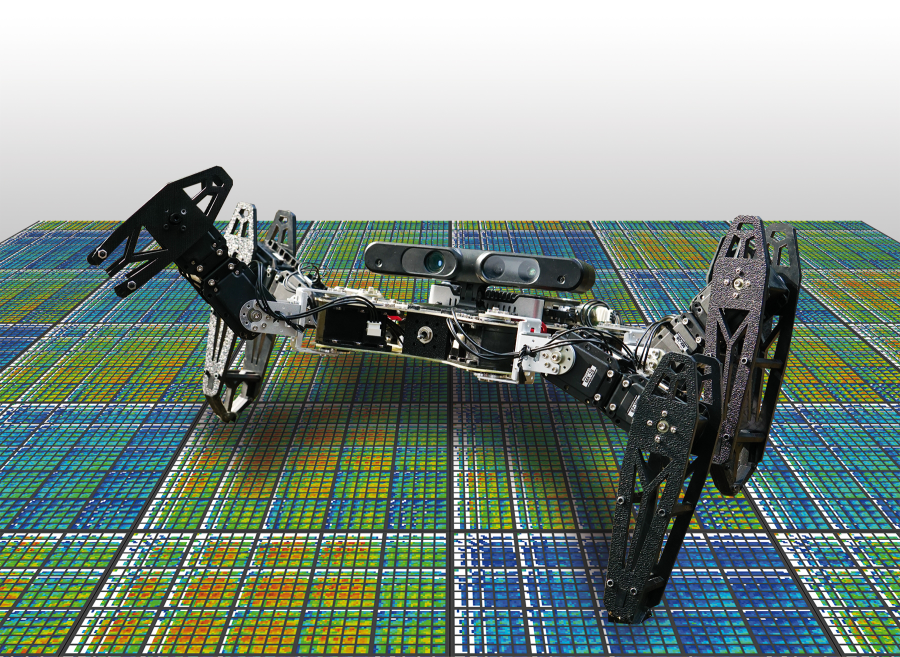}
    \caption{\textbf{Six-legged robot.} The robot uses the results of offline, simulated evolution to adapt quickly to a variety of damage conditions, such as a broken leg. Each point on the colored floor represents a different type of gait, i.e.\ a gait that uses the robot's legs in different proportions. The assumption was that that the cell in this map that required the robot to walk without using any of its legs would be impossible to fill. But, to the researchers' surprise, evolution found a way.}
    \label{fig:cullybot}
\end{figure}

Cully et al. (2015) \cite{cully2015robots} presented an algorithm that enables damaged robots to successfully adapt to damage in under two minutes. The chosen robot had six-legs, and evolution's task was to discover how to walk with broken legs or motors (Fig. \ref{fig:cullybot}). To do so, ahead of the test, the researchers coupled digital evolution with a robot simulator, to first learn a wide diversity of walking strategies. Once damaged, the robot would then use the intuitions gained from simulated evolution to quickly learn from test trials in the real world, zeroing in on a strategy that remained viable given the robot's damage.

To evolve a large diversity of gaits, the team used the MAP-Elites evolutionary algorithm~\cite{Mouret2015}, which simultaneously searches for the fittest organism over every combination of chosen dimensions of variation (i.e.\ ways that phenotypes can vary). In this case, the six dimensions of variation were the fraction of time that the foot of each leg touched the ground, a way to encourage learning diverse locomotion strategies. Thus, MAP-Elites searched for the fastest moving gait possible across every combination of how often each of the robot's six feet touched the ground. Naturally, the team thought it impossible for evolution to solve the case where all six feet touch the ground 0\% of the time, but to their surprise, it did. Scratching their heads, they viewed the video: It showed a robot that flipped onto its back and successfully walked on its elbows, with its feet in the air! (Fig. \ref{fig:cullybot2}). 
A video with examples of the different gaits MAP-Elites found, including this elbow walking gait (which is shown at the end starting at 1:49), can be viewed here: \url{https://goo.gl/9cwFtw}
\begin{figure}
    \centering
    \includegraphics[width=\textwidth]{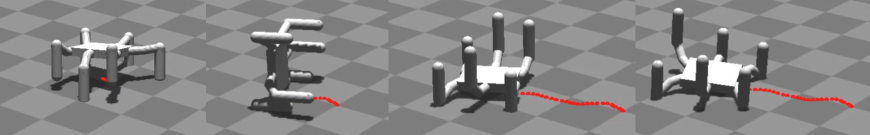}
\caption{\textbf{Elbow-walking gait.} The simulated robot, tasked with walking fast without touching any of its feet to the ground, flips over and walks on its elbows. The red line shows the center of mass of the robot over time. Note that the robot fulfills the task since the first few tenths of a second of the simulation are ignored, to focus on the gait in its limit cycle, and not the robot's initial position.}
    \label{fig:cullybot2}
\end{figure}

\paragraph{Evolution of Unconventional Communication and Information Suppression}


Mitri et al. \cite{mitri:communication,floreano:evco} applied digital evolution to groups of real and simulated robots, aiming to study the evolution of communication. The small two-wheeled robots were equipped with blue lights, which they could use as a simple channel of communication. Robots were rewarded for finding a food source while avoiding poison, both of which were represented by large red lights distinguishable only at close proximity. Over many generations of selection, all the robots evolved to find the food and avoid the poison, and under conditions that were expected to select for altruistic behavior, they also evolved to communicate the location of the food, for example by lighting up after they had reached it \cite{floreano:evco}.

However, robots also solved the problem in surprising, unanticipated ways. In some cases, when robots adapted to understand blue as a signal of food, competing robots evolved to signal blue at poison instead, evoking parallels with dishonest signaling and aggressive mimicry in nature. In other experiments that involved conditions selecting for competition between robots, authors expected that the competitive robots simply would not communicate (i.e.\ not turn on their blue light), because broadcasting the location of the food would potentially help competitors. But rather than modifying how they signaled, some robots still lit up after finding food -- but would then \emph{literally hide} the information from others by driving behind the food source (personal communication). Overall, a simple on-off light for communication revealed surprisingly rich evolutionary potential. 


\paragraph {Impossibly Compact Solutions}
To test a distributed computation platform called EC-star \cite{oreilly:ec}, Babak Hodjat implemented a multiplexer problem \cite{koza:multiplexer}, wherein the objective is to learn how to selectively forward an input signal. Interestingly, the system had evolved solutions that involved too few rules to correctly perform the task. Thinking that evolution had discovered an exploit, the impossibly small solution was tested over all possible cases. The experimenters expected this test to reveal a bug in fitness calculation. Surprisingly, all cases were validated perfectly, leaving the experimenters confused. Carefully examination of the code provided the solution: The system had exploited the logic engine's \emph{rule evaluation order} to come up with a compressed solution. In other words, evolution opportunistically offloaded some of its work into those implicit conditions.

This off-loading is similar to seminal work by Adrian Thompson in evolving real-world electronic circuits \cite{thompson:evolved}. In Thompson's experiment, an EA evolved the connectivity of a reconfigurable Field Programmable Gate Area (FPGA) chip, with the aim of producing circuits that could distinguish between a high-frequency and a lower-frequency square wave signal. After $5,000$ generations of evolution, a perfect solution was found that could discriminate between the waveforms. This was a hoped-for result, and not truly surprising in itself. However, upon investigation, the evolved circuits turned out to be extremely unconventional. The circuit had evolved to work only in the specific temperature conditions in the lab, and exploited manufacturing peculiarities of the particular FPGA chip used for evolution. Furthermore, when attempting to analyze the solution, Thompson disabled all circuit elements that were not part of the main powered circuit, assuming that disconnected elements would have no effect on behavior. However, he discovered that performance degraded after such pruning! Evolution had learned to leverage some type of subtle electromagnetic coupling, something a human designer would not have considered (or likely have known how to leverage).

\paragraph{The Fastest Route is Not Always a Straight Line}

Richard Watson and Sevan Ficici evolved the behavior of physical robots. The simple robots they built had two wheels, two motors, and two light sensors \cite{watson2002embodied,watson1999embodied}. This type of robot is well known from Braitenberg's famous book Vehicles \cite{braitenberg1986vehicles}, which argued that connecting sensor inputs to motor outputs in a particular way results in simple light-following behavior. For example, when right wheel is driven proportionally to how much light the left sensor detects and the left wheel is similarly driven by the right light sensor, the robot will move towards the light. In Watson and Ficici's case the weights of connections between the input from the light sensors and the two wheel speeds were determined by evolution. The initial question was whether Braitenberg's original solution would actually be found \cite{watson2002embodied,watson1999embodied}. 

\begin{figure}
\begin{center}
\includegraphics[height=3.5in]{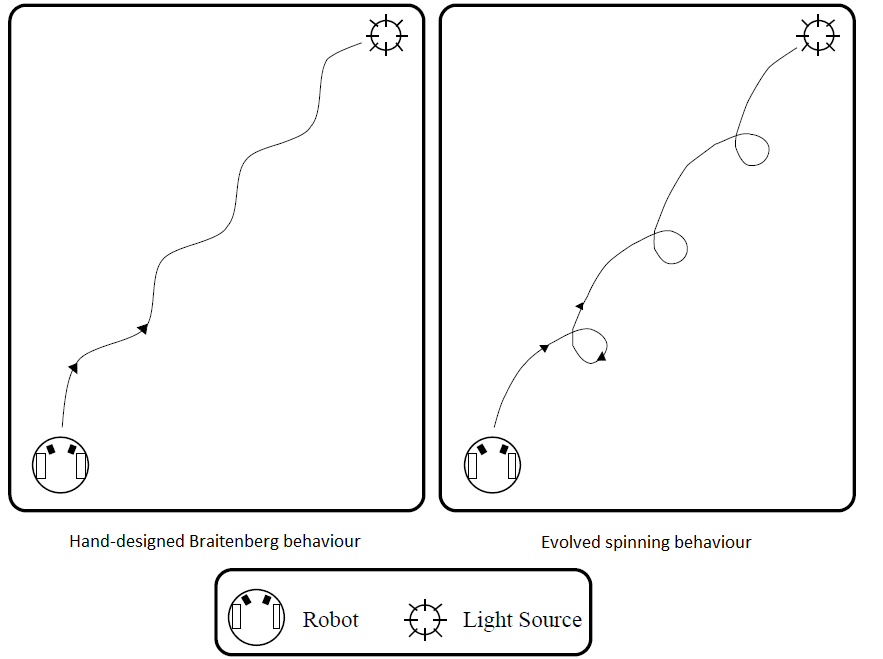}
\end{center}
\caption{\textbf{Light-seeking robot movement.} The path of the hand-coded Braitenberg-style movement (left) and evolved spinning movement (right) when moving towards a light source.}
\label{fig:spinning}
\end{figure}

While the evolved robots successfully drove towards the light source, they often did so in unusual and unintuitive ways. Some \emph{backed up} into the light while facing the dark, which was certainly an unexpected strategy. Others found the source by light-sensitive eccentric spinning, rather than the Braitenberg-style movement (Fig. \ref{fig:spinning}). It turns out that such spinning can easily be fine tuned, by tightening or loosening the curvature, to produce effective light-seeking. After some analysis the authors discovered that the portion of the genetic search space that results in spinning is \emph{extremely large}, while the classical Braitenberg solution requires delicate balance (e.g.\ the robot must execute subtle turns,  changing heading from slightly clockwise to slightly anti-clockwise, Fig. \ref{fig:spinning}) and thereby occupies a relatively tiny part of genetic space. Further, despite its apparent inefficiency, spinning remained functional even when driven at higher speeds, unlike the classical solution, which could not adjust quickly enough when run at high motor speeds. Additionally, the spinning solution was more robust to hardware differences between the individual robots, and was less likely to get stuck in corners of the arena. Thus, evolution ultimately was able to discover a novel solution that was more robust than what had initially been expected.

\paragraph{Evolving a Car without Trying}
At first glance it may seem that interactive evolution \cite{takagi:ieee01} is unlikely to surprise anyone, because of close and constant interactions with the user. And yet, in the case of Picbreeder \cite{secretan:picbreeder}, one such surprise was career-altering. Picbreeder is a platform, similar in form to the classic Blind Watchmaker experiment \cite{dawkins:blind}, where the user can evolve new designs by choosing and recombining parents, with mutations, through successive generations. The images are encoded by mathematical functions, which are invisible to the user, and may strongly constrain the direction and size of successive evolutionary steps. The surprise snuck up on one of the platform co-authors, Stanley, while evolving new images from one that resembled an alien face. As Stanley selected the parents, he suddenly noticed that the eyes of the face had descended and now looked like the wheels of a car. To his surprise, he was able to evolve a very different but visually interesting and familiar image of a car in short order from there. This quick and initially unintended transition between recognizable but dissimilar images was only the beginning of the story. The surprise inspired Stanley to conceive the novelty search algorithm \cite{lehman:ecj11}, which searches without an explicit objective (just as Stanley found the car unintentionally), selecting instead the most different, novel outcomes at each evolution step. Later formalized by Lehman and Stanley together, the now-popular algorithm owes its existence to the unexpected evolution of a car from an alien face.

\subsubsection*{Impressive Evolved Art and Design }

The anecdotes so far have focused on applications and insights related to computer science and engineering.  However, there is also a long tradition of applying digital evolution to art and design. Here we detail two such examples. What is impressive and surprising about these stories is that the outputs were not valued because they were decent attempts by computers to produce artistic creations, but were instead judged as valuable strictly on their own merits.

\paragraph {Evolving Tables and Tunes}

In the 1990s digital evolution was often applied to optimization problems, but rarely to produce novel and functional designs. Peter Bentley was interested in this challenge, but initial feedback from professional designers was dismissive and discouraging: such an approach is impossible, they said, because computers cannot invent new designs. They argued that even something as simple as a table could not be invented by evolution -- how could it possibly find the right structure from a vast sea of possibilities, and how would you specify a meaningful fitness function?

This challenge led  Bentley to create the Generic Evolutionary Design system \cite{bentley:generic}, 
which provided evolution with an expanding set of building blocks it could combine into complex configurations. 
Fitness functions were developed that rewarded separate aspects of a functional design, such as: Is the upper surface flat? Will it stand upright when supporting a mass on its upper surface? Is its mass light enough to be portable? 

\begin{figure}
\begin{center}
\includegraphics[width=\textwidth]{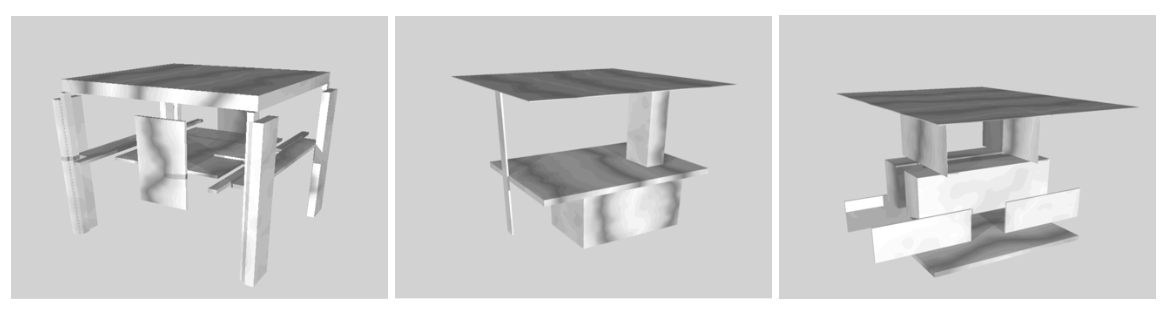}
\end{center}
\caption{\textbf{Table designs.}
Three table designs evolved using the generic evolutionary design system \cite{bentley:generic}.}
\label{fig:designs}
\end{figure}

The task put to the Generic Evolutionary Design system was to evolve a table. From random initial designs, there emerged multiple elegant designs, including a variety of different functional tables, such as the classic four-legged type, one with a small but heavy base, and one with a flat base and internal weight (the ``washing machine principle") (Fig. \ref{fig:designs}). One of the evolved tables was successfully built and has remained in functional use for nearly two decades.

In 1999 Bentley was approached by a group of musicians and developers who wanted to generate novel music through digital evolution. Dance music was popular at the time, so the team aimed to evolve novel dance tracks. They set different collections of number-one dance hits as targets, i.e.\ an evolving track would be scored higher the more it resembled the targets. The evolved results, 8-bar music samples, were evaluated by a musician who selected the ones to be combined into an overall piece, which was then professionally produced according to the evolved music score. The results were surprisingly good: The evolved tracks incorporated complex drum rhythms with interesting accompanying melodies and bass lines. Using bands such as The Prodigy as targets, digital evolution was able to produce novel dance tracks with clear stylistic resemblance. 

In 2000 the group formed a record label named J13 Records. A distribution contract was drawn up and signed with Universal Music, stipulating that the true source of the music should not be revealed (even to the distributors, because Universal Music's CEO believed that no one would want to buy computer-generated music). The companies produced several dance tracks together, some of which were then taken by other music producers and remixed. Some of the music was successful in dance clubs, with the clubgoers having no idea that key pieces of the tracks they were dancing to were the products of digital evolution.

\paragraph{An Art Museum Accepts and Displays Evolved Art Produced by Innovation Engines}

\begin{figure}
\begin{center}
\includegraphics[height=3.7in]{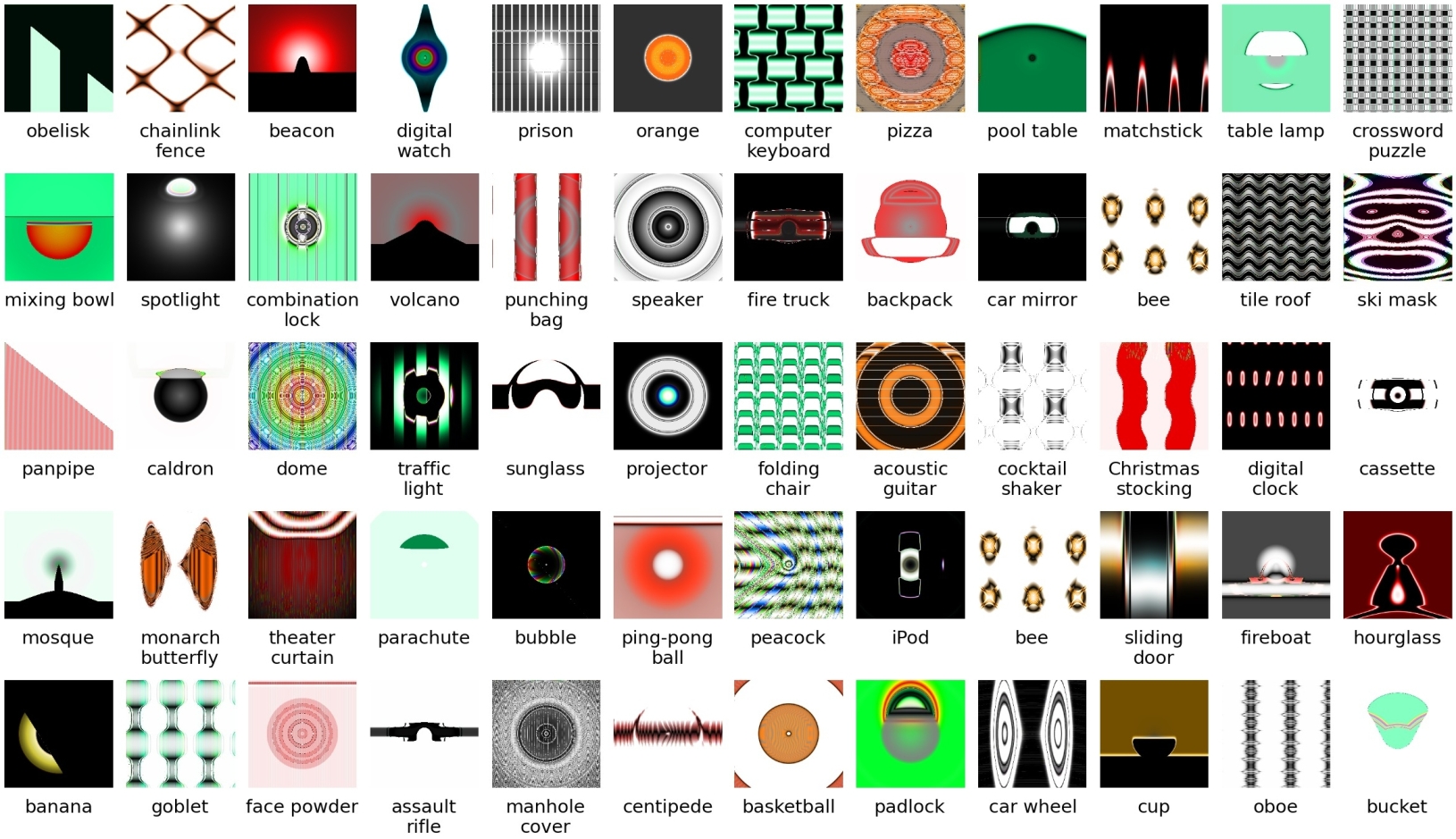}
\end{center}
\caption{\textbf{Images generated by Innovation Engines.}
A selection of images evolved via an Innovation Engine. Underneath each image is the type of image that evolution was challenged to generate.}
\label{fig:inno1}
\end{figure}

The Innovation Engine \cite{nguyen:understanding} is an algorithm that combines three keys ideas: (1) produce new innovations (i.e.\ solutions) by elaborating upon already evolved ones, (2) simultaneously evolve the population toward many different objectives (instead of a single objective as in traditional digital evolution), and (3) harness state-of-the-art deep neural networks to evaluate how interesting a new solution is. The approach successfully produced a large diversity of interesting images, many of which are recognizable as familiar objects (both to deep neural networks and human observers (Fig. \ref{fig:inno1}). Interestingly, the images span a variety of aesthetic styles, and bear resemblance to abstract ``concept art'' pieces (e.g.\ the two different images of prison cells, the beacon, and the folding chairs in Fig. \ref{fig:inno1}). Furthermore, the digital genomes of these algorithmically-produced images are quantitatively similar to the elegant, compact genomes evolved by humans on the interactive evolution website Picbreeder \cite{secretan:picbreeder}. 

\begin{figure}
\begin{center}
\includegraphics[height=2.0in]{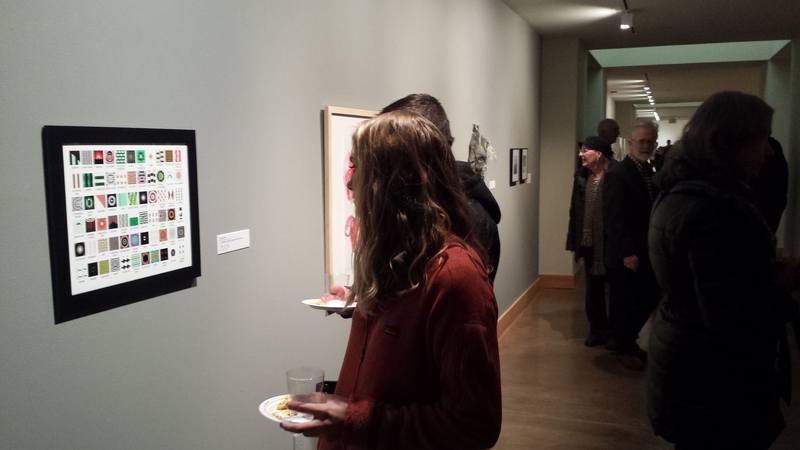}
\end{center}
\caption{\textbf{University of Wyoming art show.}
A collection of images evolved with Innovation Engines on display at the University of Wyoming Art Museum. They have also been displayed in art exhibits in galleries, fairs, and conventions in several countries around the world.}
\label{fig:inno2}
\end{figure}

To test whether the images generated by the Innovation Engine could pass for quality art, the authors submitted a selection of evolved images to a competitive art contest: the University of Wyoming's 40th Annual Juried Student Exhibition. Surprisingly, not only was the Innovation Engine piece among the 35.5\% of submissions accepted, it was also among the 21.3\% of submissions that were given an award! The piece was hung on the museum walls alongside human-made art, without visitors knowing it was evolved (Fig. \ref{fig:inno2}). 

\subsection*{Convergence with Biology}

Because it is inspired by biological evolution, digital evolution naturally shares with it the important abstract principles of selection, variation, and heritability. However, there is no guarantee that digital evolution will exhibit similar specific \emph{behaviors} and \emph{outcomes} as found in nature, because the low-level details are so divergent: mutation rates, genome sizes, how genotypes map to phenotypes, population sizes, morphology, type of interactions, and environmental complexity. Interestingly, however, this section demonstrates how in practice there often exists surprising convergence between evolution in digital and biological media.

\paragraph{Evolution of Muscles and Bones}

\begin{figure}[h]
 \begin{center}
 \includegraphics[width=\textwidth]{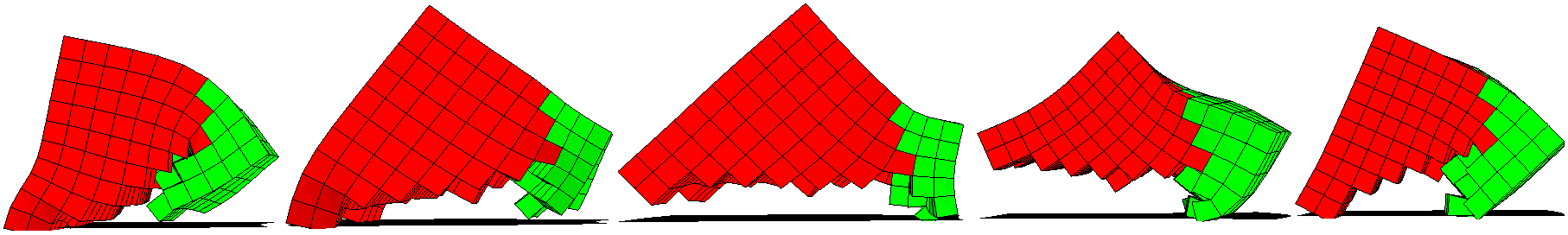}
 \includegraphics[width=\textwidth]{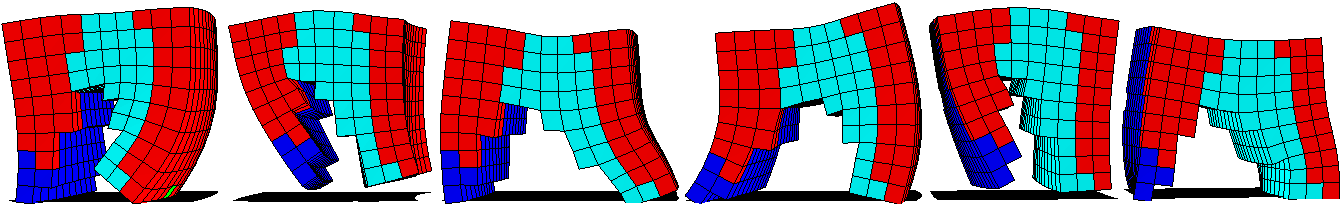}
 \end{center}
 \caption{\textbf{} A stop-motion view of a small sample of the evolved gaits from Cheney et al. \cite{cheney2013unshackling}, which produced surprisingly effective and biologically-reminiscent behaviors.  Shown here are soft robots progressing from left to right across the panel.  Colors correspond to voxel types (with red and green denoting oppositely contracting muscle groups, and dark and light blue representing stiff and soft support materials, respectively).  In the top gait, notice how evolution creates distinct regions of each muscle.  It employs these opposing muscle groups to create an inchworm-like behavior. In the bottom gait, the use of stiff (bone-like) support material allows evolution to create relatively long appendages and produce a horse-like galloping behavior. Videos of various soft robot gaits, including these two, can be found at {\url{https://youtu.be/z9ptOeByLA4?list=PL5278ezwmoxQODgYB0hWnC0-Ob09GZGe2}}.}
 \label{fig:surprisinglyGood}
 \end{figure}

In further results from Cheney et al.'s virtual creatures system \cite{cheney2013unshackling}, evolution generated  locomotion strategies unexpectedly convergent with those of biological creatures, examples of which are shown in Fig.~\ref{fig:surprisinglyGood}. The gait in the top figure is similar to the crawling of an inchworm, requiring evolution to discover from scratch the benefit of complementary (opposing) muscle groups, similar to such muscle pairs in humans, e.g.\ biceps and triceps -- and also to place them in a functional way. The gait in the bottom figure highlights digital evolution's use of a stiff bone-like material to support thinner appendages, enabling them to be longer and skinnier without sacrificing their weight-bearing potential. The end product is a gait reminiscent of a horse's gallop. 

\paragraph{Evolution of Parasitism}

In 1990, Tom Ray developed his seminal artificial life system, Tierra \cite{ray1992j}, an early instance of evolution by natural selection in a digital medium. Organisms in Tierra consist of self-replicating machine code, somewhat like computer viruses. However, unlike computer viruses, organisms in Tierra live on virtual machines explicitly designed to enable evolution (e.g.\ the instruction set was designed to be fault tolerant and evolvable). Tierra manages a population of replicating programs, killing off the oldest programs or those generating the most errors. Importantly, the operations (including copying) are faulty, meaning that replication necessarily produces mutations. Ray's hope was that Tierra would eventually create an interesting and alien tree of life in a computational universe, but he expected to spend perhaps years tinkering before anything interesting would happen; surprisingly, Tierra produced complex ecologies the very first time it ran without crashing \cite{ray1992j}.

\begin{figure}
\begin{center}
\includegraphics[height=2.0in]{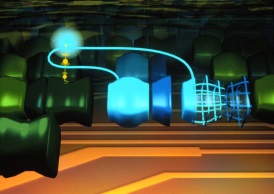}
\end{center}
\caption{\textbf{Parasites in Tierra.} A self-replicator (green, left) has code that copies the genome from parent to offspring. The parasite (blue, center) lacks the genome replicating code, and executes that code in its neighbor, copying its genome into its offspring (blue shell, right).  The blue sphere represents the parasite's CPU executing its neighbor's code. Image courtesy of Anti-Gravity Workshop.}
\label{fig:parasite}
\end{figure}

\begin{figure}
\begin{center}
\includegraphics[height=2.0in]{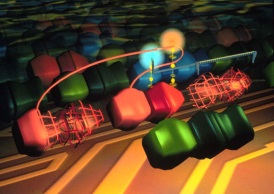}
\end{center}
\caption{\textbf{Hyper-parasites in Tierra.}
A red hyper-parasite (center), has captured a CPU (blue sphere) from a parasite, and is using it to replicate its genome into the shell on the right.  The hyper-parasite also has its own CPU (red sphere) that it is using to replicate also into the shell at the left. Image courtesy of Anti-Gravity Workshop.}

\label{fig:hyperparasite}
\end{figure}

What emerged was a series of competing adaptations between replicating organisms within the computer, i.e.\ an ongoing co-evolutionary dynamic. The surprisingly large palette of emergent behaviors included parasitism (Fig. \ref{fig:parasite}), immunity to parasitism, circumvention of immunity, hyper-parasitism (Fig. \ref{fig:hyperparasite}), obligate sociality, cheaters exploiting social cooperation, and primitive forms of sexual recombination.  All of these relied on digital templates, parts of code that provide robust addressing for JMP and CALL, the machine instructions that enable subroutines and control changes. By accessing templates not only in their own genomes, but in the genomes of others, Tierra organisms unexpectedly exploited this feature to facilitate a variety of ecological interactions.

When two individuals have complementary templates, interaction occurs. Organisms that evolved matching templates were able to execute code in neighboring organisms. They were selected for, because by outsourcing computation they reduced the size of their genome, which made replication less costly. Such organisms effectively engaged in \emph{informational} parasitism. Evolving matching templates enabled exploitation, while  non-complementary templates allowed individuals to escape exploitation. Ray termed the underlying process \emph{bit-string races}, echoing the idea of evolutionary and ecological arms-races in nature.

But the dynamics went  further than  bit-string races. Hyper-parasites stole the CPUs of parasites, exhibiting \emph{energy} parasitism. Social cooperators executed some of their own code, and some of their identical neighbor's code, to their mutual advantage. Social cheaters stole CPUs as they passed, exploiting the implicit trust between social creatures. Echoing natural evolution, a  diversity of social and ecological interactions evolved in complex ways.

\paragraph{Digital Vestigial Organs}

\begin{figure}\begin{center}\includegraphics[width=0.26\textwidth]{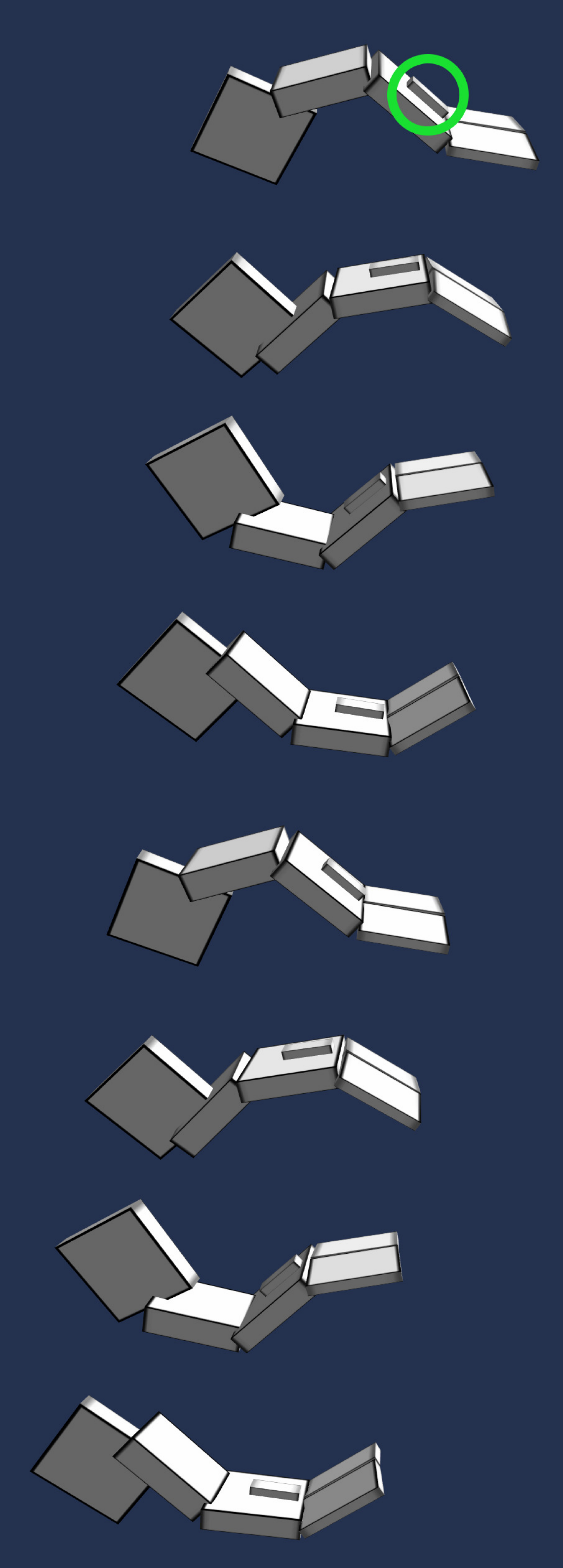}\end{center}
    \caption{\textbf{Swimming creature with an atrophied body part.} A body part that was functionally important to an ancestor of the depicted creature became atrophied over evolutionary time. Mutations within this system did not allow deleting parts entirely; as a result, evolution shrunk the part and tuned its placement to minimize its deleterious impact on swimming speed. See
    \url{https://youtu.be/JHOXzQeeUik?list=PL5278ezwmoxQODgYB0hWnC0-Ob09GZGe2} for full video.}
    \label{fig:atrophy}
\end{figure}

Virtual creatures evolved in the ERO system by Krcah~\cite{krcah08:ices} displayed a curious property: They sometimes contained small body parts whose function was not immediately obvious, yet they seemed to be carefully placed on the creature's body. It was not clear what purpose, if any, such ``decorations'' served. See Fig.~\ref{fig:atrophy} for an example of a swimming creature with an ornamental ``fin'' on top of its back.

Analysis of the ``fin'' and its evolution demonstrated that its persistence was a consequence of a specific limitation of the evolutionary algorithm: Mutation was implemented such that body parts were never entirely removed from any creature. The ``fin'' body part from Fig.~\ref{fig:atrophy} had origins as a big randomly generated block added very early in the creature's evolution. Because it could not be later removed when it started to interfere with the movements of the creature, it was instead quickly atrophied to the smallest allowed size and moved into the least obtrusive position, by a series of mutations.

\paragraph{Whole Genome Duplication in Avida}

Avida is a rich and versatile platform for digital evolution, one that has been used to study many fundamental evolutionary questions \cite{lenski2003evolutionary, adami2006digital, lenski1999genome, clune2008natural, misevic2006sexual, adami2000evolution, wilke2001evolution, chow2004adaptive,lenski2006balancing,elsberry2009cockroaches,goldsby2012task,covert2013experiments}. During the submission process for a paper on genetic architecture and sexual reproduction \cite{misevic2006sexual}, reviewers pointed out that some data was unexpectedly bi-modal: Evolution had produced two types of populations with distinct properties. After further investigation, the two types were found to be largely separable by their genome size. One group had lengths similar to the ancestral genome ($50$ instructions), while the other group had genomes about twice as long, suggestive of genome duplication events. Duplication mutations were known to be theoretically possible, but it was not obvious why or how such a sharp change in genome length had evolved. Interestingly, the Avida organisms had found an unanticipated (and unintended) mechanism to duplicate their genomes. 

Experiments in Avida typically start from a hand-coded ancestral organism, effectively a short program that consists of a series of instructions capable of self-reproduction but nothing else. The reproduction mechanism executes a loop that iteratively copies the genome instruction by instruction. The loop terminates when an ``if'' instruction detects that the last instruction in the genome has been copied. The double-length organisms resulted from an unanticipated situation, which was triggered when organisms had an odd number of instructions in their genome, and a mutation then introduced a second ``copy'' instruction into the copy loop. Because the ``if'' condition was checked only after every two ``copy'' instructions, the copying process could continue past the last instruction in the genome, ultimately copying the whole genome again. In this way, through a particular detail of the Avida reproduction mechanism, digital organisms managed to duplicate their entire genomes as sometimes also happens in biological evolution.

\paragraph{Evolving Complex Behavior May Involve Temporary Setbacks}


In a pioneering study, Richard Lenski and colleagues used the Avida digital evolution platform to test some of Darwin's hypotheses about the evolution of complex features \cite{lenski2003evolutionary}. In Avida, digital organisms can perform a wide variety of computational functions, including copying themselves, by executing instructions in their genome. The researchers were interested in the general processes by which the evolutionary process produces complex features. The team specifically focused on whether and how Avidians might evolve to perform the most complex logical function in the environment---EQU---which requires comparing two 32 bit numbers and determining whether or not they are equal.

The experiment provided several surprises about the creative power of the evolutionary process. 
The EQU function evolved in about half the replicate experimental populations, but each instance was unpredictably different, using anywhere from 17 to 43 instructions. The most surprising outcome was that the pathway that evolution followed was not always an upward climb to greater fitness, nor even a path consisting of sideways, neutral steps. Instead, in several cases, mutations along the line of descent to EQU were deleterious, some significantly so. In two cases, mutations reduced fitness by half.  Though highly deleterious themselves, these mutations produced a genetic state that allowed a subsequent beneficial mutation to complete a sequence that could preform the EQU function. This result sheds light on how complex traits can evolve by traversing rugged fitness landscapes that have fitness valleys that can be crossed to reach fitness peaks.

\paragraph{Drake's Rule} 

The Aevol digital evolution model, which belongs to the so-called ``sequence-of-nucleotides" formalism \cite{hindre_NatRevMicrob2012}, was originally developed by Carole Knibbe and Guillaume Beslon with the intent to study the evolution of modularity in gene order. However, even if some preliminary results on gene order were promising, none of them turned out statistically viable after deeper investigation, seemingly indicating that the whole project was likely to fail. However, one day in a corridor, Knibbe bumped into Laurent Duret, a renowned bioinformatician. Knibbe related her disappointing PhD advancement, saying ``We have nothing interesting; the only clear signal is that genome size apparently scales with mutation rates -- both the coding part and the non-coding part, but that's trivial, isn't it?''. Laurent disagreed, ``The non-coding part also? But that's a scoop!'' It turned out that (i) without being designed to do so, the Aevol model had spontaneously reproduced ``Drake's rule,''  stating that the size of microbial genomes scales with the inverse of their mutation rate \cite{drake_pnas1991}, and (ii) no model had predicted a scaling between the mutation rates and the non-coding size of a genome. Only the relation between the size of the \emph{coding} region of the genome and the mutation rates was theoretically expected, as a result of the error threshold effect first identified by Eigen in his quasispecies model \cite{eigen1971}. The effect on the non-coding region could be observed in Aevol because the model included chromosomal rearrangements in addition to point mutations \cite{knibbe_MolBiolEvol2007}. This random encounter opened a new research direction that ultimately led to a more general mathematical model of genome evolution, showing that indirect selection for robustness to large duplications and deletions strongly bounds genome size \cite{fischer_BulMathBiol2014}.

\paragraph{Becoming Unbreakable Rather than Better}

Many examples have described surprising ways that digital evolution optimizes its fitness function. However, early experiments aiming to evolve computer programs (a technique called genetic programming \cite{koza:gp}) revealed another kind of surprise. As evolution progressed across generations, evolved genetic programs kept becoming larger and larger, a phenomenon called ``bloat'' \cite{luke2006comparison,silva2009dynamic,langdon1998fitness}, which eventually slowed the algorithm to a crawl because it took so long to run the huge evolved programs to test their fitness. Upon closer examination, the evolved programs were found, surprisingly, to be full of ``junk code'' that could be completely removed without changing the behavior of the program at all (see also the rich literature on neutrality in digital and biological evolution \cite{galvan2011neutrality,kimura1983neutral}). 

The mystery was resolved when researchers discovered that bloat can have an evolutionary advantage by buffering against the disruptive effects of mutating and mating genetic programs \cite{Altenberg:1994:EEGP}. In early populations, creating a child program by mutating it, or replacing parts of one parent with those from another is, on average, very damaging to fitness. Once in a while, of course, the change is beneficial, which fuels evolutionary adaptation.  Additionally, some changes will by chance introduce new branches of neutral junk code that have no effect on fitness \cite{introns}. These mutants are no better at satisfying the fitness function, but their \emph{offspring} are less likely to be harmed by mutation or mating; that is because organisms with a higher percentage of neutral code have a lower chance of random changes happening to critical code. The result is that bloated individuals are more likely to produce unchanged offspring, which grants them an indirect evolutionary advantage because the average fitness of unchanged offspring is higher than the average of offspring with mutations that have affected functioning code. Over generations, programs thus grow larger and become increasingly robust to the effects of mutation and crossover.  
 
The surprise is that evolution has its own ``agenda'' distinct from the programmer's. While the programmer hopes to create an algorithm that discovers the fittest structures, evolution may instead seek the structures whose fitnesses are least disturbed by reproduction.  In some cases these two agendas may come into conflict, as they do here in two distinct ways. First, bloat causes genetic programming to consume increasing amounts of computation and memory over generations (due to rapidly growing evolved programs) \cite{langdon2000quadratic}. The more fundamental conflict is that the high fraction of junk code in these huge evolved programs requires \emph{far more generations} of evolution to adapt and improve -- because evolution learns to turn off the very thing that ultimately fuels its ability to learn: random changes that alter behavior. A similar result has been observed in digital evolution, where when evolution was given the ability to evolve its own mutation rate, with accepted wisdom being that doing so would improve the results, evolution instead short-sightedly turns off mutations entirely \cite{clune2008natural}! Just as with bloat, this behavior is beneficial in the short run because mutations tend to be harmful on average, but prevents adaptation over the long term, greatly hurting long-term performance. Interestingly, this phenomenon goes away when evolution is not forced to focus on short term fitness improvement \cite{Lehman:and:Stanley:2011:Improving}. That natural evolution could favor robustness to change has been known to biologists since the 1940s \cite{Waddington:1942,Schmalhausen:1949}. But the idea that natural selection could favor robustness over finding the fittest organisms was discovered only in the 1980s \cite{Schuster:and:Swetina:1988}, and later in digital evolution as selection for ``conservative code'' \cite{Altenberg:1994:EEGP}, neutral evolution of mutational robustness \citep{Nimwegen:Crutchfield:and:Huynen:1999,Bornberg-Bauer:and:Chan:1999:Modeling}, and ``survival of the flattest'' \cite{Wilke:Wang:etal:2001:Evolution,Schuster:and:Swetina:1988}.

\paragraph{Costly Genes Hiding from Natural Selection} 

Genes coding for cooperative behaviors --- such as public good secretion or altruistic suicide --- face very specific selection pressures that have interested researchers for decades. Their existence may seem counter-intuitive, because they bring a benefit to the population at the expense of the individuals bearing them. The Aevol system has recently been used to study cooperation \cite{misevic2012effects,frenoy2012robustness}, by giving individuals the ability to secrete a public good molecule, which benefits all digital organisms in the neighborhood. However, the public good molecule is costly for an individual  to produce, digitally mirroring the challenges facing the evolution of cooperation in biological systems. In one experiment, the researchers studied the dynamics behind the loss of such costly cooperative genes. To evolve populations that would secrete the public good molecule, the researchers first lowered the cost of the molecule. After public good secretion had evolved, the researchers continued evolution with an increased cost to study if its production would cease. Interestingly, while most populations quickly lost all their secretion genes when the cost was increased in the second stage of evolution, some populations consistently did not, even when the second stage experiments were repeated many times (starting from the same population). 

The genetic analysis of the populations in which individuals consistenly continued cooperating led to a surprising result. The secretion genes that survived the increase in cost were frequently overlapping with crucial metabolic genes, meaning that they were physically encoded using the same DNA base pairs as a metabolic gene, but using the opposite strand or another reading frame \cite{frenoy2013genetic}. As a result, it was challenging for mutations to alter secretion behavior without also destroying metabolic genes. Costly secretion genes were effectively hiding behind directly beneficial metabolic ones. There is anecdotal evidence of similar mechanisms reducing the evolutionary potential toward cheating behavior in microbes \cite{foster2004pleiotropy, nogueira2009horizontal}, but overlapping genes had never been studied in this context. Highlighting how such results may often go unappreciated and unstudied, when Frenoy, a master student at the time, manually looked at the genomes which preserved secretion despite its cost (to try to understand how they were different), he had not heard of gene overlap and thought the result was likely an uninteresting artifact of the Aevol system. Only when presenting his results during a lab meeting did his colleagues point him toward the existence of overlapping genes in nature, and the fact that selection pressures on such genetic systems are not yet fully understood.

\section*{Discussion}

A persistent misunderstanding is that digital evolution cannot meaningfully inform biological knowledge because ``it is only a simulation.'' 
As a result, it is difficult to convince biologists, other scientists, and the general public that these systems, like biological evolution, are complex, creative, and surprising. Often such disagreements occur outside of published papers, in informal conversations and responses to reviewers. During such discussions, it is common for researchers in digital evolution to relate anecdotes like those included in this paper as evidence that such algorithms indeed unleash the creativity of the Darwinian process. 
However, such arguments lack teeth when rooted in anecdotes perpetuated through oral tradition.
Thus one motivation for this paper was to collect and validate the true stories from the original scientists and collect them for posterity. 

Future work could move beyond collecting stories to directly study the prevalance of surprise among digital evolution practitoners. For example, a survey could be conducted to measure how often
researchers experience surprise (note that like the anecdotes listed here, such surveys would depend on experimenter's self-reports of surprise). To go beyond self-reports is possible but likely requires expensive interventions, e.g.\ measuring physiological signals as experimenters work and correlating physiological signals with self-reports of surprise.

A separate motivation for collecting these anecdotes is to highlight pragmatic lessons for digital evolution practioners. Foremost, a practioner should be more skeptical of their ability to correctly specify robust fitness functions, and anticipate evolution iteratively revealing such specification failures. This awareness is most important for safety-critical applications, and points to the need for careful supervision when attempting to apply digital evolution in real world systems \cite{lehman2019evolutionary}. Undesirable outcomes can result from subtle interactions between fitness functions and experimental setups, suggesting that practioners should adopt an adversarial mindset, looking for ways in which an agent could exploit seemingly innocuous design descisions. Another lesson relates to training in simulation, where evolution often exploits bugs to achieve high fitness: Practioners should regularly visualize their simulated solutions to test whether the proposed solutions are valid and reasonable. A final lesson for practioners (and beginning researchers) is to understand how the nature of academic publishing can obscure the actual (often messy) \emph{process} of human-driven research that culminates in published papers. In other words, the phenomenon described in this paper is common knowledge
to experienced digital evolution researchers, but is nowhere to be found in the academic literature. We believe the main reasons are: (1) surprise is a subjective experience and many consider such subjectivity outside the typical bounds of science, (2) the surprising phenomena were often orthogonal (or a hindrance) to the research questions that reseachers were pursuing, and (3) that the convention for academic papers (deviations from which are often punished by reviewers) is to report the final successful experiments as if they were all that were performed; the effect is to distort the human experience of applying digital evolution in practice (e.g.\ to a new domain). As a result, practioners should read papers with a critical outlook, and adjust expectations when adopting a published technique to a new domain.

Further, the ubiquity of surprising and creative outcomes in digital evolution has implications for other fields of artificial intelligence. For example, beyond their importance to digital evolution, the many examples of ``misspecified fitness functions'' in this article connect to the broader field of artificial intelligence safety: Many researchers therein are concerned with the potential for perverse outcomes from optimizing reward functions that appear sensible on their surface \cite{amodei:concrete,bostrom:superintelligence,taylor:alignment,everitt:agi}, characterized in that community as problems such as avoiding negative side effects \cite{amodei:concrete,krakovna:side}, reward hacking (also known as wire-heading) \cite{amodei:concrete,everitt:wireheading}, or more generally as AI alignment \cite{taylor:alignment}. The list compiled here provides additional concrete examples of how difficult it is to anticipate the optimal behavior created and encouraged by a particular incentive scheme. Additionally, the narratives from practitioners highlight the iterative refinement of fitness functions often necessary to produce desired results instead of surprising, unintended behaviors. Interestingly, more seasoned researchers develop better intuitions about how the creative process of evolution works, although even they sometimes still observe comical results from initial explorations in new simulations or experiments. Thus digital evolution may provide an interesting training ground for developing intuitions about incentives and optimization, to better ground theories about how to craft safer reward functions for AI agents. 

Finally, there are interesting connections between surprising results in digital evolution and the products of directed evolution in biology, wherein selection in an experimenter-controlled evolutionary process is manipulated with the hope of improving or adapting proteins or nucleic acids for practical purposes \cite{arnold:design,peisa:protein}. Echoing our ``misspecified fitness functions'' section, the first rule of directed evolution is ``you get what you select for \cite{peisa:protein}.'' Selection for exactly the property you care about in directed evolution is often difficult and time-consuming, motivating cheaper heuristics that experimenters assume will lead to the desired outcome. However, the result is often something that meets the heuristic but deviates from the ideal outcome in surprising ways \cite{zhao:directed,schmidt:directed}. In a final ironic twist, similar evolutionary arguments (applied to a higher level of biological organization) suggest that current incentive systems in science similarly produce surprising (and undesirable) byproducts \cite{smaldino:natural}. 

\section*{Conclusion}

Across a compendium of examples we have reviewed many ways in which digital evolution produces surprising and creative solutions. We have also synthesized lessons from these examples, to aid pracitioners and to communicate implications for biology and artificial intelligence more broadly. 
For every anecdote we included, there are likely others that have been already forgotten as researchers retire and new ones being created. 
The diversity and abundance of these examples suggest that surprise in digital evolution is common, rather than a rare exception,  providing evidence that evolution---whether biological or computational---is inherently creative, and should routinely be expected to surprise, delight, and even outwit us.

\section*{Acknowledgements}
We thank Elizabeth Ostrowski for a suggestion in the Digital Evolution lab at Michigan State University over a decade ago that led to the idea for this article, and for suggestions of Avida anecdotes to include. We also appreciate Tim Taylor, from whom we got the idea for crowd-sourcing the writing of this paper by an open call for material from the community in which those whose submissions were accepted would become co-authors of the paper; he successfully piloted that model for a paper we were a part of \cite{taylor2016webal}, and we adopted it here. Finally, we also thank all of those who submitted anecdotes that we were not able to include. Jeff Clune was supported by an NSF CAREER award (CAREER: 1453549).

\bibliographystyle{plos2015}
\nolinenumbers
\bibliography{bibliography/converted_to_latex.bib}

\begin{thebibliography}{100}

\bibitem{schoner:bats}
Sch{\"o}ner MG, Sch{\"o}ner CR, Simon R, Grafe TU, Puechmaille SJ, Ji LL,
  et~al.
\newblock Bats are acoustically attracted to mutualistic carnivorous plants.
\newblock Current Biology. 2015;25(14):1911--1916.

\bibitem{makarova:extreme}
Makarova KS, Aravind L, Wolf YI, Tatusov RL, Minton KW, Koonin EV, et~al.
\newblock Genome of the extremely radiation-resistant bacterium Deinococcus
  radiodurans viewed from the perspective of comparative genomics.
\newblock Microbiology and Molecular Biology Reviews. 2001;65(1):44--79.

\bibitem{strahs:bomb}
Strahs G.
\newblock Biochemistry at 1000C: Explosive secretory discharge of bombardier
  beetles (Brachinus).
\newblock Science. 1969;165(3888):61--63.

\bibitem{lefevre:invasion}
Lefevre T, Adamo SA, Biron DG, Misse D, Hughes D, Thomas F.
\newblock Invasion of the body snatchers: the diversity and evolution of
  manipulative strategies in host--parasite interactions.
\newblock Advances in Parasitology. 2009;68:45--83.

\bibitem{futuyma:natural}
Futuyma DJ.
\newblock Natural selection and adaptation.
\newblock In: The Princeton Guide to Evolution. Princeton University Press;
  2013. p. 279--301.

\bibitem{dawkins:blind}
Dawkins R.
\newblock The blind watchmaker: Why the evidence of evolution reveals a
  universe Without design.
\newblock WW Norton \& Company; 1986.

\bibitem{madigan:bacterial}
Madigan MT.
\newblock Bacterial habitats in extreme environments.
\newblock In: Journey to Diverse Microbial Worlds. Springer; 2000. p. 61--72.

\bibitem{ultrafast}
Nelson B. University of Utah biologists surprised to discover "ultrafast
  recycling" at nerve synapses; 2013.
\newblock
  \url{http://kuer.org/post/u-u-biologists-surprised-discover-ultrafast-recycling-nerve-synapses}.

\bibitem{moonlight}
{Press Release}. In Arctic winter, marine creatures migrate by the light of the
  moon; 2016.
\newblock
  \url{https://www.eurekalert.org/pub_releases/2016-01/cp-iaw123015.php}.

\bibitem{lau:censors}
Lau NC, Bartel DP.
\newblock Censors of the genome.
\newblock Scientific American. 2003;289(2):34--41.

\bibitem{fire1998potent}
Fire A, Xu S, Montgomery MK, Kostas SA, Driver SE, Mello CC.
\newblock Potent and specific genetic interference by double-stranded RNA in
  Caenorhabditis elegans.
\newblock Nature. 1998;391(6669):806--811.

\bibitem{last2016moonlight}
Last KS, Hobbs L, Berge J, Brierley AS, Cottier F.
\newblock Moonlight drives ocean-scale mass vertical migration of zooplankton
  during the Arctic winter.
\newblock Current Biology. 2016;26(2):244--251.

\bibitem{watanabe2013ultrafast}
Watanabe S, Rost BR, Camacho-P{\'e}rez M, Davis MW, S{\"o}hl-Kielczynski B,
  Rosenmund C, et~al.
\newblock Ultrafast endocytosis at mouse hippocampal synapses.
\newblock Nature. 2013;504(7479):242.

\bibitem{dob:chance}
Dobzhansky T.
\newblock Chance and creativity in evolution.
\newblock In: Studies in the Philosophy of Biology. Springer; 1974. p.
  307--338.

\bibitem{bentley:creative}
Bentley PJ.
\newblock {Is evolution creative?}
\newblock In: Proceedings of the AISB. vol.~99. Citeseer; 1999. p. 28--34.

\bibitem{dawkins:universal}
Dawkins R.
\newblock {Universal {D}arwinism}.
\newblock The Nature of Life: Classical and Contemporary Perspectives from
  Philosophy and Science. 1983;.

\bibitem{dennett:darwin}
Dennett DC.
\newblock Darwin's dangerous idea.
\newblock Simon and Schuster; 2014.

\bibitem{dejong:evolutionary}
De~Jong KA.
\newblock Evolutionary computation: A unified approach.
\newblock MIT press; 2006.

\bibitem{langton:artificial}
Langton CG.
\newblock Artificial life: An overview.
\newblock Mit Press; 1997.

\bibitem{mansanne:debug}
Mansanne F, Carrere F, Ehinger A, Schoenauer M.
\newblock Evolutionary algorithms as fitness function debuggers.
\newblock In: International Symposium on Methodologies for Intelligent Systems.
  Springer; 1999. p. 639--647.

\bibitem{pennock2000discover}
Pennock RT.
\newblock Can Darwinian mechanisms make novel discoveries?: Learning from
  discoveries made by evolving neural networks.
\newblock Foundations of Science. 2000;5(2):225--238.

\bibitem{grabowski:odometry}
Grabowski LM, Bryson DM, Dyer FC, Pennock RT, Ofria C.
\newblock A case study of the de novo evolution of a complex odometric behavior
  in digital organisms.
\newblock PLoS One. 2013;8(4):e60466.

\bibitem{darwin:origin}
Darwin C.
\newblock On the Origin of Species.
\newblock John Murray; 1859.

\bibitem{wilson:diversity}
Wilson EO.
\newblock The diversity of life.
\newblock WW Norton \& Company; 1999.

\bibitem{runco:creative}
Runco MA, Jaeger GJ.
\newblock The standard definition of creativity.
\newblock Creativity Research Journal. 2012;24(1):92--96.

\bibitem{gould:exaptation}
Gould SJ, Vrba ES.
\newblock Exaptation---a missing term in the science of form.
\newblock Paleobiology. 1982;8(01):4--15.

\bibitem{kundrat:feather}
Kundr{\'a}t M.
\newblock When did theropods become feathered?---evidence for pre-archaeopteryx
  feathery appendages.
\newblock Journal of Experimental Zoology Part B: Molecular and Developmental
  Evolution. 2004;302(4):355--364.

\bibitem{kirschner:evolvability}
Kirschner M, Gerhart J.
\newblock {Evolvability}.
\newblock Proceedings of the National Academy of Sciences of the United States
  of America. 1998;95(15):8420.

\bibitem{kouvaris2015evolution}
Kouvaris K, Clune J, Kounios L, Brede M, Watson RA.
\newblock How evolution learns to generalise: Using the principles of learning
  theory to understand the evolution of developmental organisation.
\newblock PLoS Computational Biology. 2017;13:e1005358.

\bibitem{kounios2016resolving}
Kounios L, Clune J, Kouvaris K, Wagner GP, Pavlicev M, Weinreich DM, et~al.
\newblock Resolving the paradox of evolvability with learning theory: How
  evolution learns to improve evolvability on rugged fitness landscapes.
\newblock arXiv preprint arXiv:161205955. 2016;.

\bibitem{endler:frequency}
Endler JA, Greenwood JJD.
\newblock {Frequency-dependent predation, crypsis and aposematic coloration}.
\newblock Philosophical Transactions of the Royal Society of London B,
  Biological Sciences. 1988;319(1196):505--523.

\bibitem{schluter:adaptive}
Schluter D.
\newblock The ecology of adaptive radiation.
\newblock OUP Oxford; 2000.

\bibitem{Lenski849}
Lenski RE.
\newblock Get {A} Life.
\newblock Science. 1998;280(5365):849--850.

\bibitem{dennet2002}
Dennett D.
\newblock Encyclopedia of Evolution.
\newblock In: Pagel M, editor. Encyclopedia of Evolution. Oxford Univ. Press;
  2002. p. E83--E92.

\bibitem{ofria:avida}
Ofria C, Wilke CO.
\newblock Avida: A software platform for research in computational evolutionary
  biology.
\newblock Artificial life. 2004;10(2):191--229.

\bibitem{ray:tierra}
Ray TS.
\newblock An evolutionary approach to synthetic biology: Zen and the art of
  creating life.
\newblock Artificial Life. 1993;1(1-2):179--209.

\bibitem{bedau2003artificial}
Bedau MA.
\newblock Artificial life: organization, adaptation and complexity from the
  bottom up.
\newblock Trends in cognitive sciences. 2003;7(11):505--512.

\bibitem{sims1994evolving:alife}
Sims K.
\newblock {Evolving 3D morphology and behavior by competition}.
\newblock Artificial Life. 1994;1(4):353--372.

\bibitem{bredeche2009line}
Bredeche N, Haasdijk E, Eiben A.
\newblock On-line, on-board evolution of robot controllers.
\newblock In: International Conference on Artificial Evolution (Evolution
  Artificielle). Springer; 2009. p. 110--121.

\bibitem{lehman:rr}
Lehman J, Stanley KO.
\newblock Investigating Biological Assumptions through Radical
  Reimplementation.
\newblock Artificial Life. 2014;21(1):21--46.

\bibitem{lenski2003evolutionary}
Lenski RE, Ofria C, Pennock RT, Adami C.
\newblock {The evolutionary origin of complex features}.
\newblock Nature. 2003;423(6936):139--144.

\bibitem{clune2013originModularity}
Clune J, Mouret JB, Lipson H.
\newblock The evolutionary origins of modularity.
\newblock Proceedings of the Royal Society B. 2013;280(1755):20122863.

\bibitem{Kashtan2005}
Kashtan N, Alon U.
\newblock {Spontaneous evolution of modularity and network motifs}.
\newblock Proceedings of the National Academy of Sciences.
  2005;102(39):13773--13778.

\bibitem{kashtan2007varying}
Kashtan N, Noor E, Alon U.
\newblock Varying environments can speed up evolution.
\newblock Proceedings of the National Academy of Sciences. 2007;104(34):13711.

\bibitem{lenski1999genome}
Lenski RE, Ofria C, Collier TC, Adami C.
\newblock {Genome complexity, robustness and genetic interactions in digital
  organisms}.
\newblock Nature. 1999;400(6745):661--664.

\bibitem{wagner2007road}
Wagner GP, Pavlicev M, Cheverud JM.
\newblock The road to modularity.
\newblock Nature Reviews Genetics. 2007;8(12):921--931.

\bibitem{clune2008natural}
Clune J, Misevic D, Ofria C, Lenski RE, Elena SF, Sanju{\'a}n R.
\newblock {Natural selection fails to optimize mutation rates for long-term
  adaptation on rugged fitness landscapes}.
\newblock PLoS Computational Biology. 2008;4(9):e1000187.

\bibitem{misevic2006sexual}
Misevic D, Ofria C, Lenski RE.
\newblock {Sexual reproduction reshapes the genetic architecture of digital
  organisms}.
\newblock Proceedings of the Royal Society of London B: Biological Sciences.
  2006;273(1585):457--464.

\bibitem{adami2000evolution}
Adami C, Ofria C, Collier TC.
\newblock {Evolution of biological complexity}.
\newblock Proceedings of the National Academy of Sciences of the United States
  of America. 2000;97(9):4463.

\bibitem{wilke2001evolution}
Wilke CO, Wang JL, Ofria C, Lenski RE, Adami C.
\newblock {Evolution of digital organisms at high mutation rates leads to
  survival of the flattest}.
\newblock Nature. 2001;412(6844):331--333.

\bibitem{chow2004adaptive}
Chow SS, Wilke CO, Ofria C, Lenski RE, Adami C.
\newblock {Adaptive radiation from resource competition in digital organisms}.
\newblock Science. 2004;305(5680):84.

\bibitem{cully2015robots}
Cully A, Clune J, Tarapore D, Mouret JB.
\newblock Robots that can adapt like animals.
\newblock Nature. 2015;521(7553):503--507.

\bibitem{yedid2002macroevolution}
Yedid G, Bell G.
\newblock Macroevolution simulated with autonomously replicating computer
  programs.
\newblock Nature. 2002;420(6917):810.

\bibitem{lenski2006balancing}
Lenski RE, Barrick JE, Ofria C.
\newblock Balancing robustness and evolvability.
\newblock PLoS Biology. 2006;4(12):e428.

\bibitem{goldsby2012task}
Goldsby HJ, Dornhaus A, Kerr B, Ofria C.
\newblock Task-switching costs promote the evolution of division of labor and
  shifts in individuality.
\newblock Proceedings of the National Academy of Sciences.
  2012;109(34):13686--13691.

\bibitem{covert2013experiments}
Covert AW, Lenski RE, Wilke CO, Ofria C.
\newblock Experiments on the role of deleterious mutations as stepping stones
  in adaptive evolution.
\newblock Proceedings of the National Academy of Sciences.
  2013;110(34):E3171--E3178.

\bibitem{pennock2007models}
Pennock RT.
\newblock {Models, simulations, instantiations, and evidence: the case of
  digital evolution}.
\newblock Journal of Experimental \& Theoretical Artificial Intelligence.
  2007;19(1):29--42.

\bibitem{turing:halting}
Turing AM.
\newblock On computable numbers, with an application to the
  Entscheidungsproblem.
\newblock Journal of Math. 1936;58(345-363):5.

\bibitem{langton:chaos}
Langton CG.
\newblock Computation at the edge of chaos: phase transitions and emergent
  computation.
\newblock Physica D: Nonlinear Phenomena. 1990;42(1-3):12--37.

\bibitem{flake:computational}
Flake GW.
\newblock The computational beauty of nature: Computer explorations of
  fractals, chaos, complex systems, and adaptation.
\newblock MIT press; 1998.

\bibitem{holland2000emergence}
Holland JH.
\newblock Emergence: From chaos to order.
\newblock OUP Oxford; 2000.

\bibitem{bedau1997weak}
Bedau MA.
\newblock Weak emergence.
\newblock No{\^u}s. 1997;31:375--399.

\bibitem{deutsch1998fabric}
Deutsch D.
\newblock The fabric of reality.
\newblock Penguin UK; 1998.

\bibitem{roese:hindsight}
Roese NJ, Vohs KD.
\newblock Hindsight bias.
\newblock Perspectives on Psychological Science. 2012;7(5):411--426.

\bibitem{scienceandsanity}
Korzybski A.
\newblock Science and sanity: An introduction to non-Aristotelian systems and
  general semantics.
\newblock Institute of GS; 1958.

\bibitem{campbell:law}
Campbell DT.
\newblock {Assessing the impact of planned social change}.
\newblock Evaluation and Program Planning. 1979;2(1):67--90.

\bibitem{goodhart:law}
Goodhart CA.
\newblock {Problems of monetary management: The UK experience}.
\newblock Springer; 1984.

\bibitem{sims1994evolving:compgraph}
Sims K.
\newblock {Evolving virtual creatures}.
\newblock In: Proceedings of the 21st Annual Conference on Computer Graphics
  and Interactive Techniques. ACM; 1994. p. 15--22.

\bibitem{krcah08:ices}
Krcah P.
\newblock {Towards efficient evolutionary design of autonomous robots}.
\newblock In: Evolvable Systems: From Biology to Hardware, Proceeding of the
  8th International Conference ICES. Springer; 2008. p. 153--164.

\bibitem{genprog2009}
Forrest S, Nguyen T, Weimer W, {Le Goues} C.
\newblock A genetic programming approach to automated software repair.
\newblock In: Proceedings of the Genetic and Evolutionary Computation
  Conference; 2009. p. 947--954.

\bibitem{koza:gp}
Koza JR.
\newblock Genetic programming: On the programming of computers by means of
  natural selection. vol.~1.
\newblock MIT press; 1992.

\bibitem{ssbse2013}
Weimer W.
\newblock Advances in Automated Program Repair and a Call to Arms.
\newblock In: Proceedings of Search Based Software Engineering - 5th
  International Symposium; 2013. p. 1--3.

\bibitem{schulte2010automated}
Schulte E, Forrest S, Weimer W.
\newblock Automated program repair through the evolution of assembly code.
\newblock In: Proceedings of the IEEE/ACM International Conference on Automated
  Software Engineering. ACM; 2010. p. 313--316.

\bibitem{ellefsen:neural}
Ellefsen KO, Mouret JB, Clune J.
\newblock Neural modularity helps organisms evolve to learn new skills without
  forgetting old skills.
\newblock PLoS Computational Biology. 2015;11(4):e1004128.

\bibitem{soltoggio:neuromodulation}
Soltoggio A, Bullinaria JA, Mattiussi C, Durr P, Floreano D.
\newblock Evolutionary advantages of neuromodulated plasticity in dynamic,
  reward-based scenarios.
\newblock In: Proceedings of the 11th international conference on artificial
  life (Alife XI). vol.~11. MIT Press; 2008. p. 569--576.

\bibitem{o1991monochromatic}
O'shea DC.
\newblock {Monochromatic quartet: A search for the global optimum}.
\newblock In: International Lens Design Conference. International Society for
  Optics and Photonics; 1991. p. 548--554.

\bibitem{gagne2008human}
Gagn{\'e} C, Beaulieu J, Parizeau M, Thibault S.
\newblock {Human-competitive lens system design with evolution strategies}.
\newblock Applied Soft Computing. 2008;8(4):1439--1452.

\bibitem{cheney2013unshackling}
Cheney N, MacCurdy R, Clune J, Lipson H.
\newblock {Unshackling evolution: evolving soft robots with multiple materials
  and a powerful generative encoding}.
\newblock In: Proceedings of the Genetic and Evolutionary Computation
  Conference. ACM; 2013. p. 167--174.

\bibitem{moriarty:discovering}
Moriarty DE, Miikkulainen R.
\newblock {Discovering complex {O}thello strategies through evolutionary neural
  networks}.
\newblock Connection Science. 1995;7(3):195--209.

\bibitem{moriarty:ec97}
Moriarty DE, Miikkulainen R.
\newblock {Forming neural networks through efficient and adaptive
  co-evolution}.
\newblock Evolutionary Computation. 1997;5(4):373--399.

\bibitem{Feldt1998GPSW}
Feldt R.
\newblock {Generating diverse software versions with genetic programming: An
  experimental study}.
\newblock IEE Proceedings - Software Engineering. 1998;145(6):228--236.

\bibitem{Feldt1999GPExplorative}
Feldt R.
\newblock {Genetic programming as an explorative tool in early software
  development phases}.
\newblock In: Proceedings of the 1st International Workshop on Soft Computing
  Applied to Software Engineering; 1999. p. 11--20.

\bibitem{Feldt2002Thesis}
Feldt R.
\newblock Biomimetic software engineering techniques for dependability.
\newblock Department of Computer Engineering, Chalmers University of
  Technology. Gothenburg, Sweden; 2002.

\bibitem{Harman:2012:SSE:2379776.2379787}
Harman M, Mansouri SA, Zhang Y.
\newblock {Search-based software engineering: Trends, techniques and
  applications}.
\newblock ACM Comput Surv. 2012;45(1):11:1--11:61.

\bibitem{stanley:ieeetec05}
Stanley KO, Bryant BD, Miikkulainen R.
\newblock Real-time neuroevolution in the {NERO} video game.
\newblock IEEE Transactions on Evolutionary Computation. 2005;9(6):653--668.

\bibitem{mnih2015human}
Mnih V, Kavukcuoglu K, Silver D, Rusu AA, Veness J, Bellemare MG, et~al.
\newblock Human-level control through deep reinforcement learning.
\newblock Nature. 2015;518(7540):529.

\bibitem{salimans2017evolution}
Salimans T, Ho J, Chen X, Sutskever I.
\newblock Evolution strategies as a scalable alternative to reinforcement
  learning.
\newblock arXiv preprint arXiv:170303864. 2017;.

\bibitem{mnih2016asynchronous}
Mnih V, Badia AP, Mirza M, Graves A, Lillicrap T, Harley T, et~al.
\newblock Asynchronous methods for deep reinforcement learning.
\newblock In: International Conference on Machine Learning; 2016. p.
  1928--1937.

\bibitem{such2017deep}
Such FP, Madhavan V, Conti E, Lehman J, Stanley KO, Clune J.
\newblock Deep Neuroevolution: Genetic Algorithms Are a Competitive Alternative
  for Training Deep Neural Networks for Reinforcement Learning.
\newblock arXiv preprint arXiv:171206567. 2017;.

\bibitem{chrabaszcz2018back}
Chrabaszcz P, Loshchilov I, Hutter F.
\newblock Back to Basics: Benchmarking Canonical Evolution Strategies for
  Playing Atari.
\newblock arXiv preprint arXiv:180208842. 2018;.

\bibitem{rechenberg1973evolutionsstrategie}
Rechenberg I.
\newblock Evolutionsstrategie--Optimierung technisher Systeme nach Prinzipien
  der biologischen Evolution.
\newblock Frommann-Holzboog; 1973.

\bibitem{james2019verge}
Vincent J.
\newblock A video game-playing AI beat Q*bert in a way no one's ever seen
  before.
\newblock The Verge. 2018;.

\bibitem{Ecarlat2015}
Ecarlat P, Cully A, Maestre C, Doncieux S.
\newblock {Learning a high diversity of object manipulations through an
  evolutionary-based babbling}.
\newblock In: Proceedings of Learning Objects Affordances Workshop at IROS
  2016; 2015. p. 1--2.

\bibitem{Mouret2015}
Mouret JB, Clune J.
\newblock {Illuminating search spaces by mapping elites}.
\newblock arXiv preprint arXiv:150404909. 2015; p. 1--15.

\bibitem{moriarty:sab96}
Moriarty DE, Miikkulainen R.
\newblock {Evolving obstacle avoidance behavior in a robot arm}.
\newblock In: From Animals to Animats 4: Proceedings of the Fourth
  International Conference on Simulation of Adaptive Behavior. Cambridge, MA:
  MIT Press; 1996. p. 468--475.

\bibitem{vandersmagt:neurocomputing94}
{van der Smagt} P.
\newblock {{S}imderella: {A} robot simulator for neuro-controller design}.
\newblock Neurocomputing. 1994;6(2):281--285.

\bibitem{mitri:communication}
Mitri S, Floreano D, Keller L.
\newblock The evolution of information suppression in communicating robots with
  conflicting interests.
\newblock Proceedings of the National Academy of Sciences.
  2009;106(37):15786--15790.

\bibitem{floreano:evco}
Floreano D, Mitri S, Magnenat S, Keller L.
\newblock Evolutionary conditions for the emergence of communication in robots.
\newblock Current Biology. 2007;17(6):514--519.

\bibitem{oreilly:ec}
O'Reilly UM, Wagy M, Hodjat B.
\newblock Ec-star: A massive-scale, hub and spoke, distributed genetic
  programming system.
\newblock In: Genetic Programming Theory and Practice X. Springer; 2013. p.
  73--85.

\bibitem{koza:multiplexer}
Koza JR.
\newblock A hierarchical approach to learning the Boolean multiplexer function.
\newblock In: Foundations of Genetic Algorithms, Vol. 1. Elsevier; 1990. p.
  171--192.

\bibitem{thompson:evolved}
Thompson A.
\newblock An evolved circuit, intrinsic in silicon, entwined with physics.
\newblock In: International Conference on Evolvable Systems. Springer; 1996. p.
  390--405.

\bibitem{watson2002embodied}
Watson RA, Ficici SG, Pollack JB.
\newblock Embodied evolution: Distributing an evolutionary algorithm in a
  population of robots.
\newblock Robotics and Autonomous Systems. 2002;39(1):1--18.

\bibitem{watson1999embodied}
Watson RA, Ficiei S, Pollack JB.
\newblock Embodied evolution: Embodying an evolutionary algorithm in a
  population of robots.
\newblock In: Proceedings of the 1999 Congress on Evolutionary Computation.
  IEEE; 1999. p. 335--342.

\bibitem{braitenberg1986vehicles}
Braitenberg V.
\newblock Vehicles: Experiments in synthetic psychology.
\newblock MIT press; 1986.

\bibitem{takagi:ieee01}
Takagi H.
\newblock Interactive evolutionary computation: {F}usion of the capacities of
  {EC} optimization and human evaluation.
\newblock Proceedings of the IEEE. 2001;89(9):1275--1296.

\bibitem{secretan:picbreeder}
Secretan J, Beato N, D'Ambrosio DB, Rodriguez A, Campbell A, Folsom-Kovarik JT,
  et~al.
\newblock Picbreeder: A case study in collaborative evolutionary exploration of
  design space.
\newblock Evolutionary Computation. 2011;19(3):373--403.

\bibitem{lehman:ecj11}
Lehman J, Stanley KO.
\newblock Abandoning objectives: Evolution through the search for novelty
  alone.
\newblock Evolutionary Computation. 2011;19(2):189--223.

\bibitem{bentley:generic}
Bentley PJ.
\newblock Generic evolutionary design of solid objects using a genetic
  algorithm.
\newblock The University of Huddersfield; 1996.

\bibitem{nguyen:understanding}
Nguyen A, Yosinski J, Clune J.
\newblock Understanding innovation engines: Automated creativity and improved
  stochastic optimization via deep learning.
\newblock Evolutionary Computation. 2016;24(3).

\bibitem{ray1992j}
Ray T.
\newblock {J'ai jou{\'e} {\'a} Dieu et cr{\'e}{\'e} la vie dans mon
  ordinateur}.
\newblock Le Temps strat{\'e}gique. 1992;47:68--81.

\bibitem{adami2006digital}
Adami C.
\newblock Digital genetics: Unravelling the genetic basis of evolution.
\newblock Nature Reviews Genetics. 2006;7(2):109--118.

\bibitem{elsberry2009cockroaches}
Elsberry WR, Grabowski LM, Ofria C, Pennock RT.
\newblock Cockroaches, drunkards, and climbers: Modeling the evolution of
  simple movement strategies using digital organisms.
\newblock In: Proceedings of IEEE Symposium on Artificial Life. IEEE; 2009. p.
  92--99.

\bibitem{hindre_NatRevMicrob2012}
Hindr{\'e} T, Knibbe C, Beslon G, Schneider D.
\newblock {New insights into bacterial adaptation through in vivo and in silico
  experimental evolution}.
\newblock Nature Reviews Microbiology. 2012;10(5):352--365.

\bibitem{drake_pnas1991}
Drake JW.
\newblock {A constant rate of spontaneous mutation in DNA-based microbes}.
\newblock Proceedings of the National Academy of Sciences.
  1991;88(16):7160--7164.

\bibitem{eigen1971}
Eigen M.
\newblock {Self-organization of matter and the evolution of biological
  macromolecules.}
\newblock Die Naturwissenschaften. 1971;58(10):465--523.

\bibitem{knibbe_MolBiolEvol2007}
Knibbe C, Coulon A, Mazet O, Fayard JM, Beslon G.
\newblock {A long-term evolutionary pressure on the amount of non-coding DNA}.
\newblock Molecular Biology and Evolution. 2007;24(10):2344--2353.

\bibitem{fischer_BulMathBiol2014}
Fischer S, Bernard S, Beslon G, Knibbe C.
\newblock {A model for genome size evolution}.
\newblock Bulletin of Mathematical Biology. 2014;76(9):2249--2291.

\bibitem{luke2006comparison}
Luke S, Panait L.
\newblock A comparison of bloat control methods for genetic programming.
\newblock Evolutionary Computation. 2006;14(3):309--344.

\bibitem{silva2009dynamic}
Silva S, Costa E.
\newblock Dynamic limits for bloat control in genetic programming and a review
  of past and current bloat theories.
\newblock Genetic Programming and Evolvable Machines. 2009;10(2):141--179.

\bibitem{langdon1998fitness}
Langdon WB, Poli R.
\newblock Fitness causes bloat.
\newblock In: Soft Computing in Engineering Design and Manufacturing. Springer;
  1998. p. 13--22.

\bibitem{galvan2011neutrality}
Galv{\'a}n-L{\'o}pez E, Poli R, Kattan A, O’Neill M, Brabazon A.
\newblock Neutrality in evolutionary algorithms… What do we know?
\newblock Evolving Systems. 2011;2(3):145--163.

\bibitem{kimura1983neutral}
Kimura M.
\newblock The neutral theory of molecular evolution.
\newblock Cambridge University Press; 1983.

\bibitem{Altenberg:1994:EEGP}
Altenberg L.
\newblock The evolution of evolvability in genetic programming.
\newblock In: Advances in Genetic Programming, Vol. 1. MIT Press; 1994. p.
  47--74.

\bibitem{introns}
Nordin P, Francone F, Banzhaf W.
\newblock Explicitly Defined Introns and Destructive Crossover in Genetic
  Programming.
\newblock In: Advances in Genetic Programming, Vol. 2. MIT Press; 1996. p.
  111--134.

\bibitem{langdon2000quadratic}
Langdon WB.
\newblock Quadratic bloat in genetic programming.
\newblock In: Proceedings of the 2nd Annual Conference on Genetic and
  Evolutionary Computation. Morgan Kaufmann Publishers Inc.; 2000. p. 451--458.

\bibitem{Lehman:and:Stanley:2011:Improving}
Lehman J, Stanley KO.
\newblock Improving evolvability through novelty search and self-adaptation.
\newblock In: Evolutionary Computation (CEC), 2011 IEEE Congress on. IEEE;
  2011. p. 2693--2700.

\bibitem{Waddington:1942}
Waddington CH.
\newblock Canalization of development and the inheritance of acquired
  characters.
\newblock Nature. 1942;150(3811):563--565.

\bibitem{Schmalhausen:1949}
Schmalhausen II.
\newblock Factors of Evolution: The Theory of Stabilizing Selection.
\newblock Chicago: University of Chicago Press; 1949.

\bibitem{Schuster:and:Swetina:1988}
Schuster P, Swetina J.
\newblock Stationary mutant distributions and evolutionary optimization.
\newblock Bulletin of Mathematical Biology. 1988;50(6):635--660.

\bibitem{Nimwegen:Crutchfield:and:Huynen:1999}
Nimwegen E, Crutchfield JP, Huynen M.
\newblock Neutral Evolution of Mutational Robustness.
\newblock Proceedings of the National Academy of Sciences USA.
  1999;96(17):9716--9720.

\bibitem{Bornberg-Bauer:and:Chan:1999:Modeling}
Bornberg-Bauer E, Chan HS.
\newblock Modeling evolutionary landscapes: mutational stability, topology, and
  superfunnels in sequence space.
\newblock Proceedings of the National Academy of Sciences.
  1999;96(19):10689--10694.

\bibitem{Wilke:Wang:etal:2001:Evolution}
Wilke CO, Wang JL, Ofria C, Lenski RE, Adami C.
\newblock Evolution of digital organisms at high mutation rates leads to
  survival of the flattest.
\newblock Nature. 2001;412(6844):331--333.

\bibitem{misevic2012effects}
Misevic D, Fr{\'e}noy A, Parsons DP, Taddei F.
\newblock Effects of public good properties on the evolution of cooperation.
\newblock In: Artificial Life Conference Proceedings 12. MIT Press; 2012. p.
  218--225.

\bibitem{frenoy2012robustness}
Fr{\'e}noy A, Taddei F, Misevic D.
\newblock Robustness and evolvability of cooperation.
\newblock In: Artificial Life Conference Proceedings 12. MIT Press; 2012. p.
  53--58.

\bibitem{frenoy2013genetic}
Fr{\'e}noy A, Taddei F, Misevic D.
\newblock Genetic architecture promotes the evolution and maintenance of
  cooperation.
\newblock PLoS Computational Biology. 2013;9(11):e1003339.

\bibitem{foster2004pleiotropy}
Foster KR, Shaulsky G, Strassmann JE, Queller DC, Thompson CR.
\newblock Pleiotropy as a mechanism to stabilize cooperation.
\newblock Nature. 2004;431(7009):693--696.

\bibitem{nogueira2009horizontal}
Nogueira T, Rankin DJ, Touchon M, Taddei F, Brown SP, Rocha EP.
\newblock Horizontal gene transfer of the secretome drives the evolution of
  bacterial cooperation and virulence.
\newblock Current Biology. 2009;19(20):1683--1691.

\bibitem{lehman2019evolutionary}
Lehman J.
\newblock Evolutionary Computation and AI Safety: Research Problems Impeding
  Routine and Safe Real-world Application of Evolution.
\newblock arXiv preprint arXiv:190610189. 2019;.

\bibitem{amodei:concrete}
Amodei D, Olah C, Steinhardt J, Christiano P, Schulman J, Man{\'e} D.
\newblock Concrete problems in AI safety.
\newblock arXiv preprint arXiv:160606565. 2016;.

\bibitem{bostrom:superintelligence}
Bostrom N.
\newblock Superintelligence: Paths, dangers, strategies.
\newblock OUP Oxford; 2014.

\bibitem{taylor:alignment}
Taylor J, Yudkowsky E, LaVictoire P, Critch A.
\newblock Alignment for advanced machine learning systems.
\newblock Technical Report, Machine Intelligence Research Institute. 2016;.

\bibitem{everitt:agi}
Everitt T, Lea G, Hutter M.
\newblock AGI Safety Literature Review.
\newblock arXiv preprint arXiv:180501109. 2018;.

\bibitem{krakovna:side}
Krakovna V, Orseau L, Martic M, Legg S.
\newblock Measuring and avoiding side effects using relative reachability.
\newblock arXiv preprint arXiv:180601186. 2018;.

\bibitem{everitt:wireheading}
Everitt T, Hutter M.
\newblock Avoiding wireheading with value reinforcement learning.
\newblock In: Artificial General Intelligence. Springer; 2016. p. 12--22.

\bibitem{arnold:design}
Arnold FH.
\newblock Design by directed evolution.
\newblock Accounts of Chemical Research. 1998;31(3):125--131.

\bibitem{peisa:protein}
Peisajovich SG, Tawfik DS.
\newblock Protein engineers turned evolutionists.
\newblock Nature Methods. 2007;4(12):991--994.

\bibitem{zhao:directed}
Zhao H, Arnold FH.
\newblock Combinatorial protein design: Strategies for screening protein
  libraries.
\newblock Current Opinion in Structural Biology. 1997;7(4):480--485.

\bibitem{schmidt:directed}
Schmidt-Dannert C, Arnold FH.
\newblock Directed evolution of industrial enzymes.
\newblock Trends in Biotechnology. 1999;17(4):135--136.

\bibitem{smaldino:natural}
Smaldino PE, McElreath R.
\newblock The natural selection of bad science.
\newblock Royal Society Open Science. 2016;3(9):160384.

\bibitem{taylor2016webal}
Taylor T, Auerbach JE, Bongard J, Clune J, Hickinbotham S, Ofria C, et~al.
\newblock WebAL Comes of Age: A review of the first 21 years of Artificial Life
  on the Web.
\newblock Artificial life. 2016;22(3):364--407.

\end{thebibliography}

\end{document}